%% file: main.tex
\documentclass[a4paper,conference]{IEEEtran}
\ifCLASSINFOpdf
\else
\fi

\usepackage{balance} 
\usepackage{times}
\usepackage[T1]{fontenc}
\usepackage[latin9]{inputenc}
\usepackage{verbatim}
\usepackage{amsmath,amsfonts}
\usepackage{graphicx}
\usepackage{microtype}
\usepackage{subfig}
\usepackage{algorithm,algpseudocode}
\usepackage[inline]{enumitem}
\usepackage[american]{babel}

\usepackage{wrapfig,lipsum,booktabs}

\begin{document}
%
\title{Fast Model-based Policy Search for Universal Policy Networks}

\author{
\IEEEauthorblockN{Buddhika Laknath Semage*, Thommen George Karimpanal, Santu Rana and Svetha Venkatesh}
\IEEEauthorblockA{Applied Artificial Intelligence Institute\\
Deakin University\\
Geelong, Australia\\
*Email: bsemage@deakin.edu.au}}


%


\maketitle

\begin{abstract}
\input{abstract.tex}
\end{abstract}


%
\IEEEpeerreviewmaketitle

\section{Introduction\label{sec:intro}}

\input{intro_lk.tex}



\section{Methodology\label{sec:method}}

\input{method.tex}

\section{Experiments\label{sec:Experiments}}

\input{exp.tex}

\section{Related Work\label{sec:bg}}

\input{bg.tex}

\section{Conclusion\label{sec:conclusion}}

\input{conclude.tex}



{\small{}\bibliographystyle{IEEEtran}
\bibliography{physics}
}{\small\par}

%



\pagebreak\clearpage{}
	
\section{Appendix\label{sec:supp}}

\input{supp.tex}

\end{document}

%% file: abstract.tex
Adapting an agent's behaviour to new environments has been one
of the primary focus areas of physics based reinforcement learning.
Although recent approaches such as universal policy networks partially
address this issue by enabling the storage of multiple policies trained in simulation on a wide range of dynamic/latent factors, efficiently
identifying the most appropriate policy for a given environment
remains a challenge. In this work, we propose a Gaussian Process-based
prior learned in simulation, that captures the likely performance
of a policy when transferred to a previously unseen environment. We integrate
this prior with a Bayesian Optimisation-based policy search process to improve the efficiency of 
identifying the most appropriate policy from the universal policy network. We empirically evaluate our approach in a
range of continuous and discrete control environments, and show that
it outperforms other competing baselines.

%% file: intro_lk.tex
Over the last decade, deep reinforcement learning (RL) agents have
achieved a number of milestones such as matching and exceeding human level performances in games such as Go \cite{silver2016mastering} and  Atari \cite{mnih2015human}. However, their poor sample
efficiency is a major impediment to real-world applications where
it is impractical to collect millions of training samples \cite{mnih2015human}.
For physics-based tasks such as robotics, numerical simulators based on known
laws of physics can be used as a cheap surrogate of the real-world.
However, such simulators often have many simulation parameters,
usually representing real-world latent factors such as mass, friction etc.,
which need to be estimated (i.e. system identification) before the simulator can be used
for learning.


A common approach to this problem is estimating parameters from trajectory observations (sequence of states for a fixed duration of time) in the real world by minimising the difference between simulated and real world trajectories \cite{using-inaccurate-models,farchy2013humanoid, Zheng2018-bk,Chang2016-pp,DBLP:conf/rss/Xu0ZTS19}. However, these works are focussed on grounding the system as a whole, i.e., estimating all the simulation parameters without considering the downstream task. In most situations, not all parameters are important for a given task. For example, in basketball, bounciness of the ball is likely to be more relevant in comparison to surface friction, whereas in ten-pin bowling, the reverse is true. Thus, in situations where the task specification is available, spending samples to estimate all the parameters accurately (i.e. `full grounding') may be completely unnecessary.


Domain Randomisation \cite{8202133,DBLP:conf/rss/SadeghiL17} proposes
to find a task-specific but environment-independent robust policy
across a wide range of values of the simulation parameters. 
But in practice this approach has been shown to produce overly 
conservative policies, especially if dynamics vary considerably
across the full range of parameters \cite{data-driven-dr}. A more useful line of research
that utilises task-specific knowledge to only `partially ground' the
system relies on training a Universal Policy Network (UPN) \cite{yu2019sim}
modelled through a large deep model and trained across a range of parameter values.
These models are capable of producing environment specific
policies when conditioned through the corresponding environment parameters.
The system identification is performed by trying out different sets
of parameter values, rolling out the corresponding UPN policies in
the real environment (i.e., transfer performance) and finding the 
parameter set that maximises the reward \cite{lazaric2012transfer}. This can be
performed by any efficient optimiser such as Bayesian optimisation
(BO) \cite{frazier2018tutorial}. Partial grounding happens automatically
as the optimiser progressively identifies the influence of individual
parameters and use that knowledge in the optimisation. \emph{However, current 
BO-based approaches start with no prior knowledge of the task and thus
lack the power to exploit any intrinsic relations between different
policies for improved sample-efficiency.} 

With this motivation, we first propose a `self-play'-like approach where
we evaluate the performance of the UPN policies
individually across different values of parameters and build prior
knowledge on the inter-policy relationships, all within the simulation environment. 
We discovered that (a) in many tasks, some policies are universally good, i.e., using that policy
across a range of parameters often produces similarly good rewards,
and (b) there exist policies that perform well only for a narrow range
of parameters. Such knowledge, especially the existence of universally 
good policies can make policy search easier and efficient. We then encode such knowledge in BO
via a transfer learning method to further accelerate the BO-based policy 
search approach \cite{joy2019flexible}.  Specifically, we discretize
the parameter values, and at each sampled point we gather 
the set of simulated performance values by rolling out its UPN policy at different 
parameter regimes in the simulator. We then create a synthetic 
performance observation at that sampled point by assuming the observation to be Gaussian
with mean and variance calculated from the set of the simulated 
performance values. If the samples do not follow a Gaussian distribution,
the observation is dropped. This ensures that the prior observations
are consistent with the Gaussian process (GP) that is used as the 
main probabilistic model for BO. Next, we use the transfer learning
method in \cite{joy2019flexible} to treat these synthetic observations
as source observations and create a GP that inherits
the rich knowledge from this source data. Using the method
enables our approach to be no worse in convergence than standard BO without any
prior observations. We observe that such acceleration
in BO produces much faster policy identification than the vanilla
BO approach. We evaluate our approach in three MuJoCo
tasks\cite{openai-gym} and two contextual bandit scenarios, PHYRE \cite{Bakhtin2019-dq} 
and our own Basketball environment, and observe
that our method can produce significant improvements over state-of-the-art
methods. In short, our contributions are: 
\begin{enumerate}
\item Proposing a conceptual framework using inter-policy relationship for
accelerating policy search on Universal Policy Network based simulator-in-the-loop
learning. 
\item Deriving a transfer learning-based Bayesian optimisation method to
exploit the inter-policy relationship in a principled way. 
\item Demonstration of the the benefit of our approach on a set of physics
tasks from both contextual bandit and classic RL settings. 
\end{enumerate}
Although we have proposed our approach for physics based RL settings, we believe it can be adapted to any parameterisable environment where the parameters are not necessarily physical, but still influences the dynamics
of the environment e.g., skill of another agent in the same environment,
user preferences for recommendations, etc. Thus, we see a broader impact
of our work in areas beyond physics based RL settings.

%% file: method.tex
In this work, we propose a simulator-in-the-loop learning algorithm that 
requires minimal interaction with the real world for policy learning in an RL setting.


To achieve this we adopt an RL approach commonly
referred to as Universal Policy Networks (UPN), that makes it practical
to store and evaluate a large collection of policies for different latent factors
(e.g. mass, friction) without having to re-train the agents for each
latent factor combination. We train the UPN as an RL agent, for which
we formalise the learning problem as a Markov Decision Process (MDP). The MDP is represented as a tuple
of $<\mathcal{S},\mathcal{A},\mathcal{T},\theta,\mathcal{R}>$, where $\mathcal{S}$ is the state space, $\mathcal{A}$ is the
action space and $\theta$ is the set of latent factors. Furthermore,
$\mathcal{T}:\mathcal{S}\times \mathcal{A}\times\theta\rightarrow \mathcal{S}$ is the transition function,
and $\mathcal{R}:\mathcal{S}\times \mathcal{A}\rightarrow\mathbb{R}$ is the reward function. UPN's state
is constructed by combining the task's observable state (e.g. object positions and velocity)
and true latent factors (e.g. friction), which makes it possible for an RL
algorithm to learn policies over different latent factors and often seamlessly generalise them to
unseen latent factors if the policy space is smooth with respect to the latent parameters. 

In our problem setup, we consider a real world environment $\psi_r$, and assume the availability of a corresponding simulated environment $\psi$.  We assume these environments are governed by transition functions $\mathcal{T}_r$ and $\mathcal{T}_s$ respectively, that differ in their respective latent factors $\Theta$ and $\theta$ ($\Theta,\theta \in \mathbb{R}^d$). $\Theta$, associated with the real environment is given by $\Theta=\theta_r + \epsilon$ where 
$\theta_r \in \mathbb{R}^d$ represents some unknown, but modelable, fixed latent factor value and $\epsilon \in \mathbb{R}^d$ represents some unmodellable dynamics (e.g. air resistance).
The transition function $\mathcal{T}_s$ of the simulated environment $\psi$ can be controlled through its latent factors $\theta$.

Our aim is to maximise performance on a real world task set $W$ with a fixed reward function, by efficiently learning a policy $\pi$ through minimal interactions with the real-world. To achieve this objective, we 
aim to learn a sufficiently good value of $\theta$ to the extent that $\pi$ trained on dynamics $\mathcal{T}_s$ would return a relatively high performance when evaluated on $W$ with dynamics $\mathcal{T}_r$. 
For this purpose, we evaluate $T$ policies $\pi_{n=1,..,T}$ fetched (i.e. conditioned) 
from the UPN at different $\theta$, on $W$. We refer to this process as policy search where the sequence of $\pi_n$ is constructed by Bayesian Optimisation (BO) for discovering the optimal policy within a small value of $T$. To perform policy search with a reduced number of interactions with the real-world, we integrate a prior constructed by
evaluating the goodness of $\pi$ over a range of $\theta$. We call this prior \emph{Policy prior}.

The overall design of our approach consists of two components. 1)
a UPN based Policy Search; and 2) \emph{Policy prior }incorporated
into the BO based latent factor estimation process (Figure \ref{fig:policy-network}).
These components are discussed in the following sections:

\begin{figure}
\begin{centering}
\includegraphics[width=0.25\paperheight]{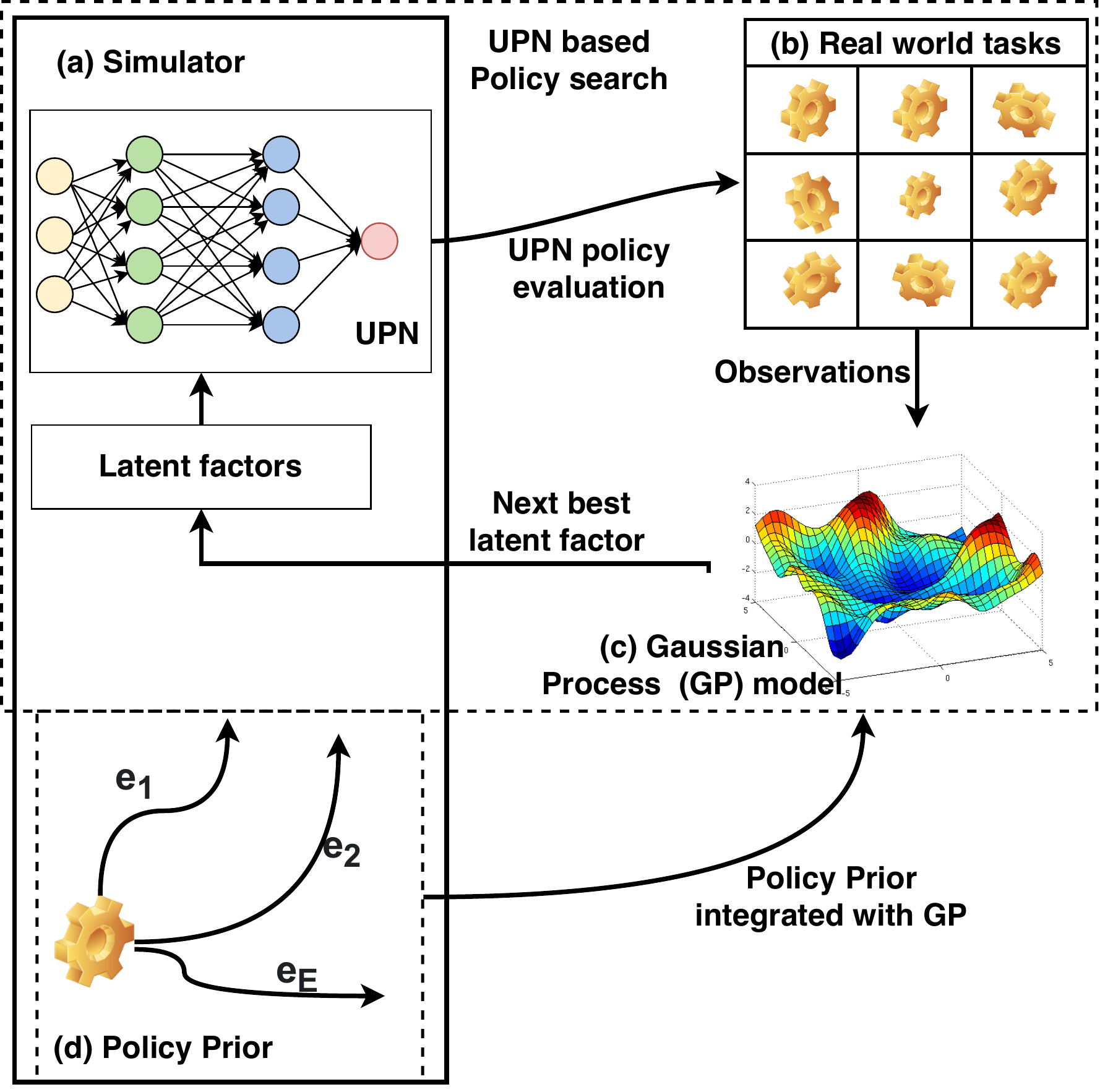}
\par\end{centering}
\centering{}\caption{\label{fig:policy-network} A sample efficient online system identification
workflow using a universal policy network (UPN) based policy search
and \emph{Policy Prior}. (a) The latent factor conditioned policy from
UPN is (b) transferred to a set of real-world tasks, and (c) evaluations are
fed as observations to a GP model. GP is used as part of BO based
policy search. (d) To improve GP model's
sample efficiency, we integrate \emph{Policy Prior }which is the simulated
prior generated by rolling out a task policy in $e_1,..,e_E$ environment settings 
with different latent factors.}
\end{figure}

\subsection{UPN Based Policy Search\label{subsec:Latent-Factor-Estimation}}

To formulate the UPN, we augment the task's state $s_{t}\in S$ as
$s_{t}=[o_{t},\theta]$, where $o_{t}$ is the task's observable state
at time step $t$, and $\theta$
is the set of latent factors for the given episode (we assume, $\theta$ is
fixed for a given episode throughout this study). Even though UPNs
have been implemented with policy gradient algorithms for mainly continuous
action tasks \cite{up-net,yu2018policy}, in this study we also design
a DQN based UPN for discrete action tasks such as bandit tasks. 
However, to examine the performance of \emph{Policy
prior} for continuous action tasks, we adopt the existing policy gradient
based UPN. When training the UPN, we only use simulation, making it
inexpensive and practical.

Once we obtain a trained UPN, when conditioned by the latent factor
estimate $\theta,$ it will produce a policy $\pi(s_{t}=[o_{t},\theta])$
for the time step $t$. We then evaluate this policy on the real-world
task set $W$, and measure its performance (episodic return) $G_{W}(\pi)$. We use this measure as the
optimisation objective for BO to find $\theta^{*}$ such that,

\begin{equation}
\theta^{*}=\underset{\theta}{argmax}\:G_{W}(\pi(s_{t}))\;\mid s_{t}=[o_{t},\theta]\label{eq:5}
\end{equation}

\subsection{\label{subsec:Building-the-gp-prior} Policy Prior}



To build the \emph{Policy prior}, we first uniformly sample $N$ latent
factors and retrieve policies $\pi_{n=1,..,N}$ for each corresponding
$X_{n=1,..N}$ latent factors by concatenating $X_n$ with task observation state at
the given time-step and using this augmented state to fetch a policy from the UPN (i.e., conditioning). 
Secondly, we define an environment setting $e$ as a simulation setting with some
fixed latent factor configuration (i.e., $\psi_\theta$ for some $\theta$), and we select a set of uniformly sampled 
environment settings $E$ with $e \in E$ from defined bounds for the
latent factors. To get a measure of the behaviour of $\pi_{n=1..N}$
policies in these $E$ settings, we evaluate each of $\pi_{n}$
on all $E$ simulated settings and record their performances. Subsequently, 
we calculate the distribution statistics,
mean $\mu_{n=1,..,N}$ and variance $\sigma_{n=1,..,N}^{2}$ of jump
starts for $E$ evaluations of each $\pi_{n}$. 

\begin{algorithm}[t] 
\caption{Policy Prior Formation $(policy\_prior)$} 
\label{alg:physics_prior} \hspace*{\algorithmicindent}
\textbf{Input:} \begin{algorithmic}[1] 
\State $U$ - a simulator trained Universal Policy Network
\State $\xi \in W$ - a set of tasks
\State $\psi$ - a simulated environment
\State $N$ - Number of observations to sample
\State $e \in E$ - Environment settings with different ground truth latent values
\State $B_l$ - Bounds for latent factors $l$, $l \in \{1..L\}$
\State $\gamma$ - Minimum threshold of P-value for Gaussianity of sampled points

\State \textbf{Output:} $X_{1..N}$ latent factor samples, $\mu_{1,..N}$, $\sigma^2_{1,..N}$ mean and variance of UPN policies evaluated in $E$ environments

\State //Sample latent values
\For{$l \gets 1$ to $L$}
\State $X_l \gets$ Generate $N$ uniform distributed observations in range $B_l$
\EndFor

\State //Evaluate sampled latent values across environments
\For{$n \gets 1$ to $N$}
	\State $\pi_n \gets U$ conditioned policy at $X_{n}$
	\For{$e \gets 1$ to $E$}
		\State $Y_{n, e} \gets$ $\pi_n$ evaluated in setting $e$ of simulator $\psi$ on task set $W$
	\EndFor
	\State $\mu_n \gets mean(Y_{n,1..E})$
	\State $\sigma^2_n \gets variance(Y_{n,1..E})$
	\State //Call Kolmogorov-Smirnov test function \cite{ktest} to calculate the p-value of $X_n$ sample for Gaussianity check. Sec. \ref{subsec:gaussianity-check}
	\State $p\_val_n = kstest(Y_{n,e=1..E}, \mu_n, \sigma^2_n)$ 
	\If {$p\_val_n < \gamma$}
		\State Remove $X_n$  
	\EndIf
\EndFor

\end{algorithmic} 
\end{algorithm}

\begin{algorithm}[t] 
\caption{Physics aware BO Policy Search} 
\label{alg:policy_search} \hspace*{\algorithmicindent}
\textbf{Input:} \begin{algorithmic}[1] 
\State $T$ - Number of BO iterations to run, $k$ - GP kernel, $\phi$ - GP initial hyperparameters, $\alpha$ - acquisition function, $\psi_r$ - real world, $W$ - task set to evaluate

\State \textbf{Output:} $x^*_{i=1..T}$, $y^*_{i=1..T}$ BO estimated new latent factors and corresponding real-world jump start performances\\
\State //Get Policy prior latent samples, mean and variance
\State $X_{1..N}$, $\mu_{1..N}, \sigma^2_{1..N} \gets policy\_prior$ 
\State //Construct covariance matrix using kernel k
\State $K \gets k(X_{1..N}, X_{1..N};\phi)$
\State //Add Policy prior variance as noise variance to Kernel matrix diagonal entries + $\epsilon$ sim-to-real noise. Eq. \ref{eqn:gp_prior}.
\For{$i \gets 1$ to $N$} $K_{i,i} \gets K_{i,i} + \sigma^2_i + \epsilon$ \EndFor
\State //Build the Policy prior integrated Gaussian process
\State $GP_{p} \gets GP(\mu_{1..N}, K)$

\State //Run the BO policy search for T iterations
\For{$i \gets 1$ to $T$}
\State //acquisition function provides the next most likely latent factors to improve the optimisation objective
\State $x^*_i \gets \alpha(GP_{p})$ 
\State // UPN conditioned policy at the latent factors suggested by acquisition function
\State $\pi_i \gets$ $U$ conditioned by $x^*_i$
\State $y^*_i \gets$ $\pi_i$ evaluated on task set $W$ in real-world $\psi_r$
\State $GP_{p} \gets GP_{p} \cup (x^*_i, y^*_i)$
\EndFor
\end{algorithmic} 
\end{algorithm}

\subsubsection{\label{subsec:gaussianity-check}Checking for Gaussianity of Prior}
In the BO process, a Gaussian process is used as the model of the
function. As per the modelling assumption of the Gaussian process,
each observation should strictly follow a Gaussian distribution. Any
mismatch between this assumption and reality will make the resultant
model inconsistent and BO can get affected. Therefore, as an additional
step, we verify whether the prior is distributed normally using Kolmogorov-Smirnov
Goodness-of-Fit Test \cite{NIST-handbook}. For this purpose, we
iterate through all sampled latent values in the \emph{Policy} \emph{prior}
to verify whether their p-values are within a defined threshold, and
if they are not, the respective points are dropped from the prior.

\subsubsection{\label{subsec:bo-transfer} Knowledge Transfer with BO}
To use these synthetic experiences 
as prior observations in the GP and then use it in the BO process we use the 
transfer BO algorithm of \cite{joy2019flexible}. In \cite{joy2019flexible}, the transfer 
was done from a set of source observations to a target optimization problem. 
In our case, the source observations are the synthetic observations 
and the target observations come from the given real-world environment. 
With that, the synthetic observations now can be included just 
as a set of extra observations with added noise. The noise allows 
the GP to adjust for the expected difference of performance in the real-world, 
compared to the simulation world. 
This prior augmented GP's observation data at time $t$ of the optimization can be denoted 
as $D_t$ = $\{X_n, \mu_n, \sigma^2_n\}_{n=1}^N \cup \{X_i^*, y_i^*\}_{i=1}^t$, where $\{X_i^*, y_i^*\}$
are real-world observations. This results in the kernel matrix form as:


\begin{equation}
\label{eqn:gp_prior}
  K_{*} = K + \begin{bmatrix}\begin{bmatrix}\sigma_{1} &  & 0\\
 					& ..\\
					0 &  & \sigma_{N}
					\end{bmatrix}_{N\times N} \\
 					& \sigma_{r}^{2}I_{t\times t}
				\end{bmatrix}
\end{equation}

where $K$ is the covariance matrix for $\{X_{n=1}^N \cup X^*_{i=1..t}\}$, and $\sigma^2_{r} \in \mathbb{R}$ is the noise variance for real-world observations.

The posterior is computed based on this $K_*$ and $D_t$. We choose this transfer learning BO algorithm because 
of its favourable convergence property that shows the convergence is no worse than a 
standard BO without prior observations (3.1, Theorem 2 of \cite{joy2019flexible}).


The process of building \emph{Policy prior} is detailed in Algorithm \ref{alg:physics_prior},
while the overall \emph{Policy prior} integrated BO policy search
is explained in Algorithm \ref{alg:policy_search}.

%% file: exp.tex
\begin{figure}
\begin{centering}
\subfloat[\label{fig:Bowling}Bowling - keep the green and blue balls in contact]{\includegraphics[width=0.15\paperwidth]{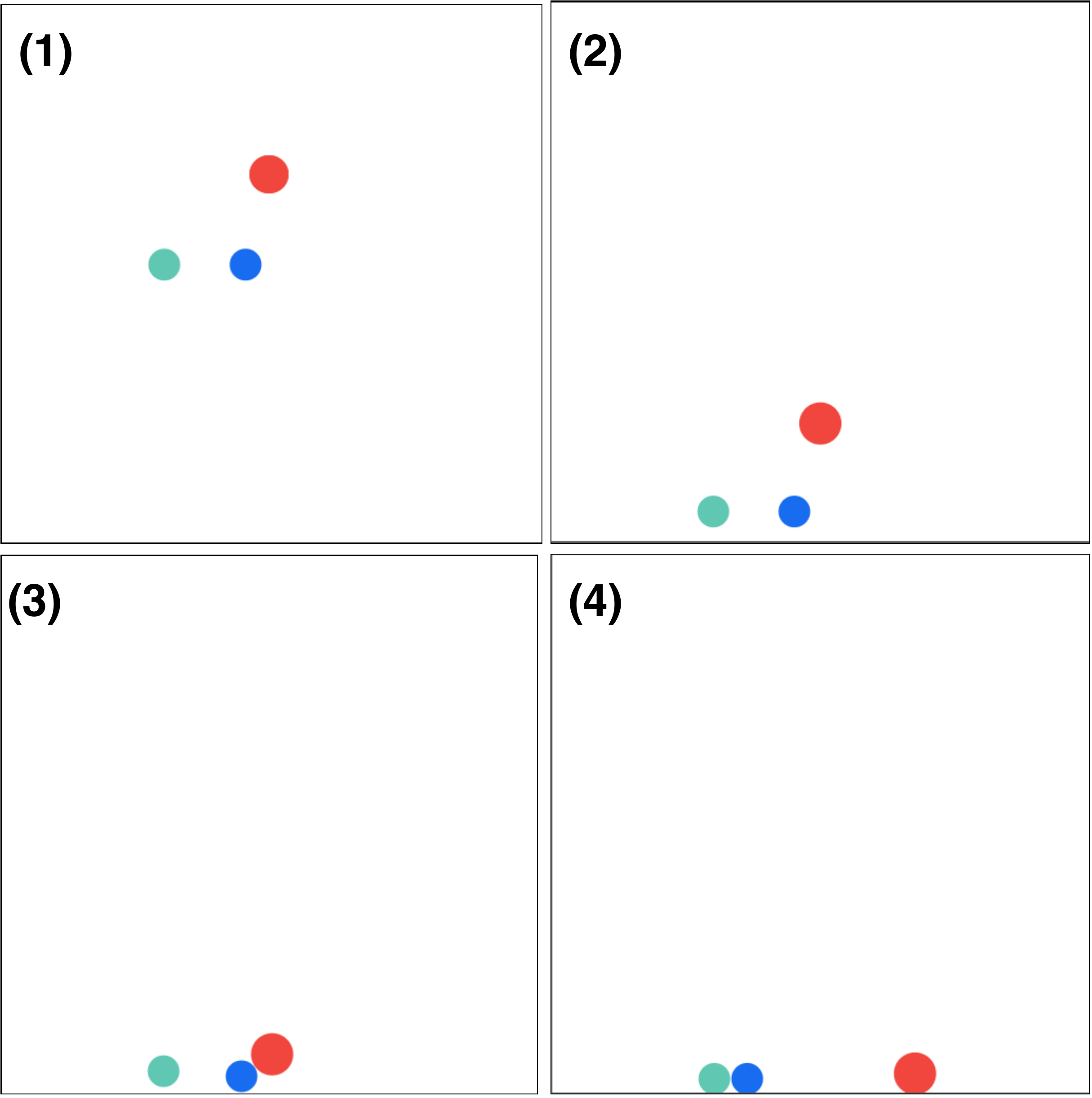}

}\qquad{}\subfloat[\label{fig:Basketball}Basketball - throw the ball into the basket
using a plank]{\centering{}\includegraphics[width=0.15\paperwidth]{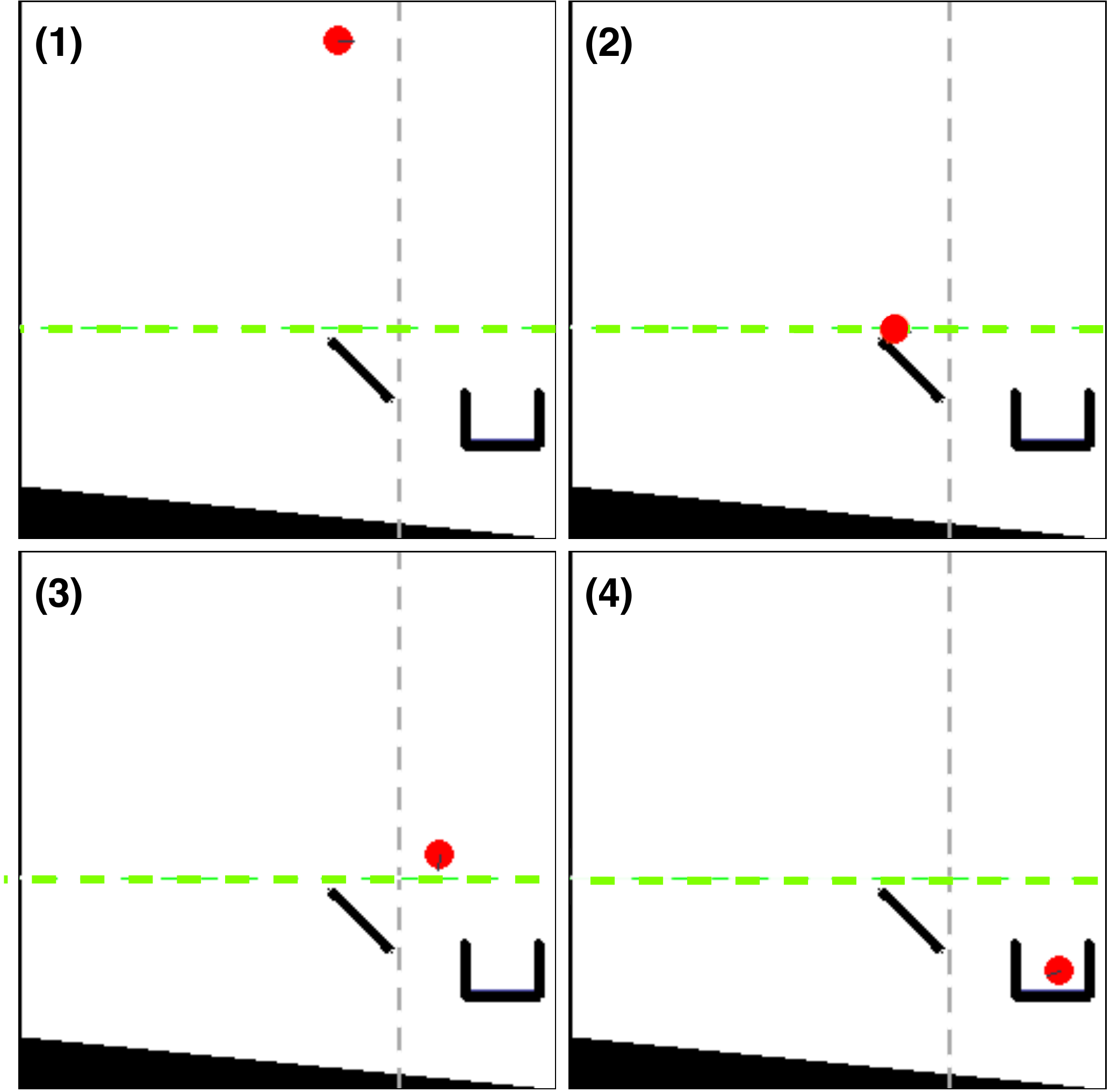}}
\par\end{centering}
\begin{centering}
\caption{Evaluated contextual bandit tasks. a) Green and blue balls are initialised
with different locations and sizes. An agent can choose the position
(x, y coordinates) and the radius of the red ball with the aim of
keeping the blue and green balls in contact for 3 seconds. b) The
ball and the plank (fixed at 45 degree angle) need to be positioned
so that the ball bounces into the basket. The plank position is constrained
to be below the horizontal green line and to the left of the vertical
gray line. \label{fig:contextual-bandit}}
\par\end{centering}
\end{figure}

\subsection{Experiment setup}

To demonstrate our approach, we consider a range
of physics-based tasks - two contextual bandit tasks and three continuous
control RL tasks. For continuous control tasks, we adopt three MuJoCo
tasks: Half Cheetah-v1, Hopper-v1 and Walker2D-v1 \cite{openai-gym}, implemented using MuJuCo physics simulator \cite{todorov2012mujoco}. 
For contextual bandit tasks with discrete actions,
we utilise two tasks named \emph{Bowling}
and \emph{Basketball} (Fig. \ref{fig:contextual-bandit}) implemented
with the Pymunk physics simulator \cite{pymunk}. \emph{Bowling} is derived
from 0000 PHYRE \cite{Bakhtin2019-dq} template, and consists of 3 balls, where the goal is to keep two of the balls (blue and green) in contact
by controlling the position and size of the red ball (i.e. action)
(Fig. \ref{fig:Bowling}) and dropping it on the blue ball. In \emph{Basketball}, the aim is to bounce a ball into a basket using an angled plank (Fig. \ref{fig:Basketball}).

For the purpose of this study, we emulate the real-world through the same simulator but with an added damping factor (i.e., a drag force).
To evaluate the performance of policies in contextual
bandit tasks, we use the \emph{AUCCESS} score from PHYRE, which is a metric designed to measure both the sample efficiency, as well as the reward obtained. It is defined
as follows: given attempts $k\in\lbrace1,..,100\rbrace$
at solving a given task, weights $w_{k}$ is determined as $w_{k}=log(k+1)-log(k)$,
then the \emph{AUCCESS }score is $\sum_{k}w_{k}.s_{k}/\sum_{k}w_{k}$
where $s_{k}$ is the success percentage at $k$ attempts \cite[Sec. 3.2]{Bakhtin2019-dq}.

\noindent\textbf{Baselines:} The baselines considered in all our experiments include learning without a prior (\emph{No Prior}), and a  Domain Randomisation (\emph{DR}) baseline, where an agent learns a policy on randomised latent factors. Additionally, for MuJoCo tasks we compare against \emph{MAML}\cite{finn2017model} as a meta-learning baseline where a set of randomised latent factors is used as tasks to train a fast adaptable meta-objective. For the contextual bandit tasks, we also consider an \emph{Estimated} baseline, where the latent factor values are estimated by minimising the differences in the trajectories of the real and simulation environments. To compensate for real-world interactions of the policy search, non-BO baselines (\emph{DR}, \emph{MAML} and \emph{Estimated}) are additionally trained for an equivalent number of interactions as the policy search in the real-world.

\subsection{UPN Based Policy Search with Bayesian Optimisation}

For the contextual bandit setting, we use a DQN based UPN implementation to find policies for our contextual
bandit tasks. To train UPNs for continuous
action tasks, we adopt the UPN implemention\footnote{https://github.com/VincentYu68/policy\_transfer}
of \cite{yu2018policy}, on which a policy is learnt using Proximal Policy Optimization the (PPO)
algorithm \cite{DBLP:journals/corr/SchulmanWDRK17}. We train our
three MuJoCo tasks for $2\times10^{8}$ steps on this UPN implementation
and build separate UPNs for each task, on which we conduct our policy
search process to find a good policy for a given environment.


Given a trained UPN and an environment with unknown latent factors,
we execute 20 iterations of Bayesian optimisation (BO) to search
for a good policy for the environment. To provide a stable optimisation
objective for BO in the MuJoCo environments, we use the sum of rewards
over a window of 500 interaction steps as the optimisation
objective. Furthermore, to isolate the effects of latent factors, we fix all other influencing
factors during a BO trial by initialising each iteration
from a fixed initial state. After BO policy search, the best estimated latent factor value is used to condition the UPN to fetch the policy, which is transferred to the real environment to evaluate the jump start performance (i.e., cumulative reward of the transferred policy) over 100 random episodes with a 500 step horizon.

For contextual bandit tasks, we maintain three separate folds of task sets 
for training, validation and testing purposes. For these tasks, we conduct BO search on the DQN based
UPN using the validation fold of datasets, with the objective set to maximise the reward over tasks in the fold. For each
BO evaluation, the total number of interactions is the number of actions attempted $\times$ number of tasks in the validation fold.




\subsection{Evaluation of the \emph{Policy Prior}}

\begin{figure*}
\centering{}
\subfloat[Hopper-2Dim]{\includegraphics[width=0.23\paperwidth]{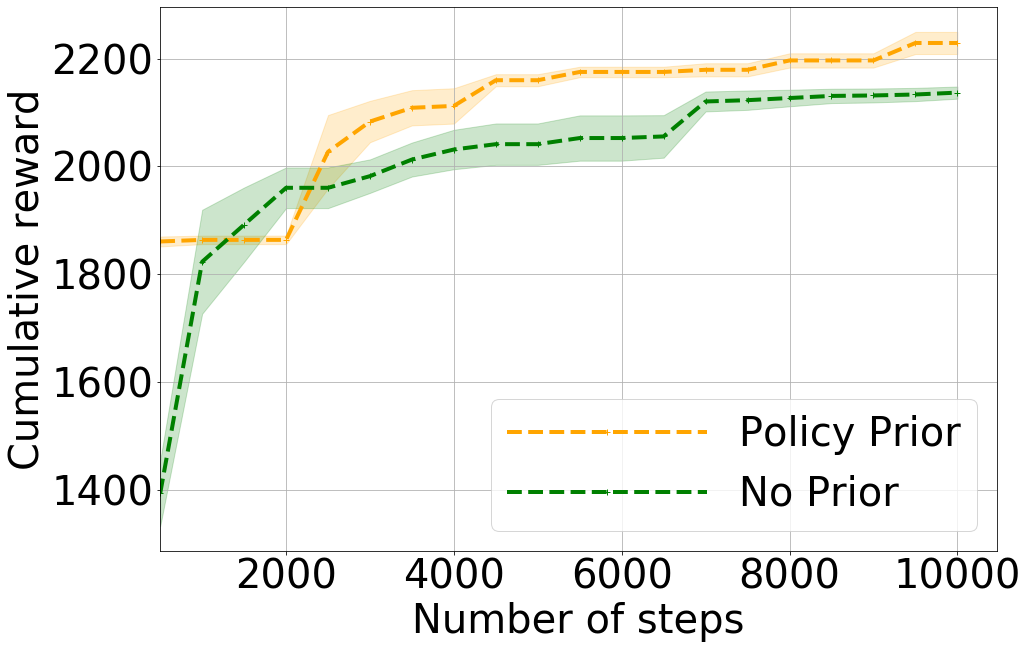}
}
\subfloat[Walker2D-5Dim]{\includegraphics[width=0.23\paperwidth]{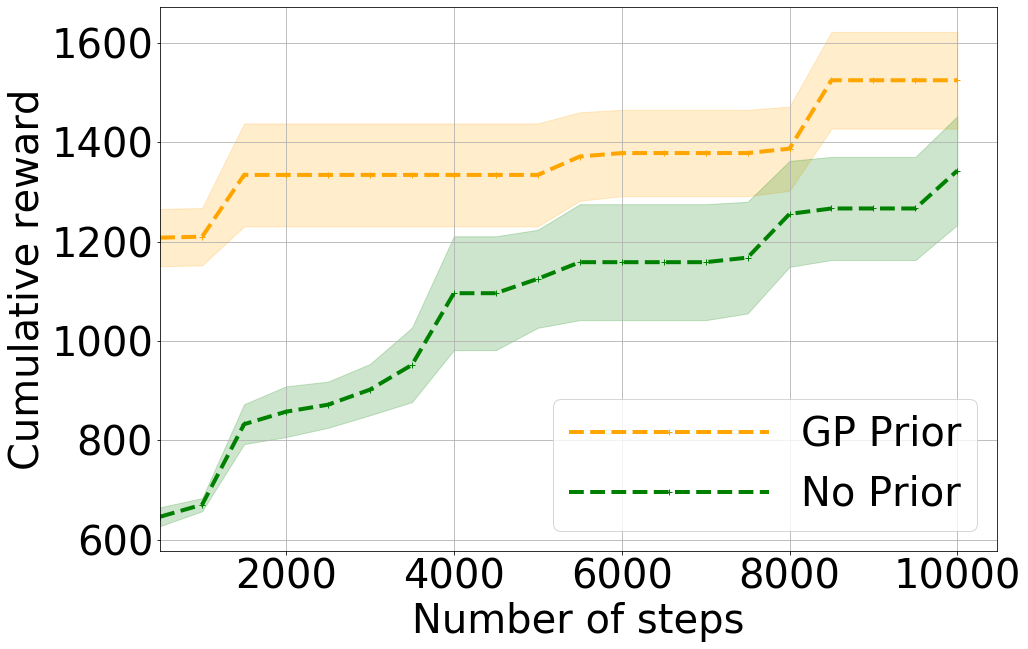}
}
\subfloat[Half-Cheetah-5Dim]{\includegraphics[width=0.23\paperwidth]{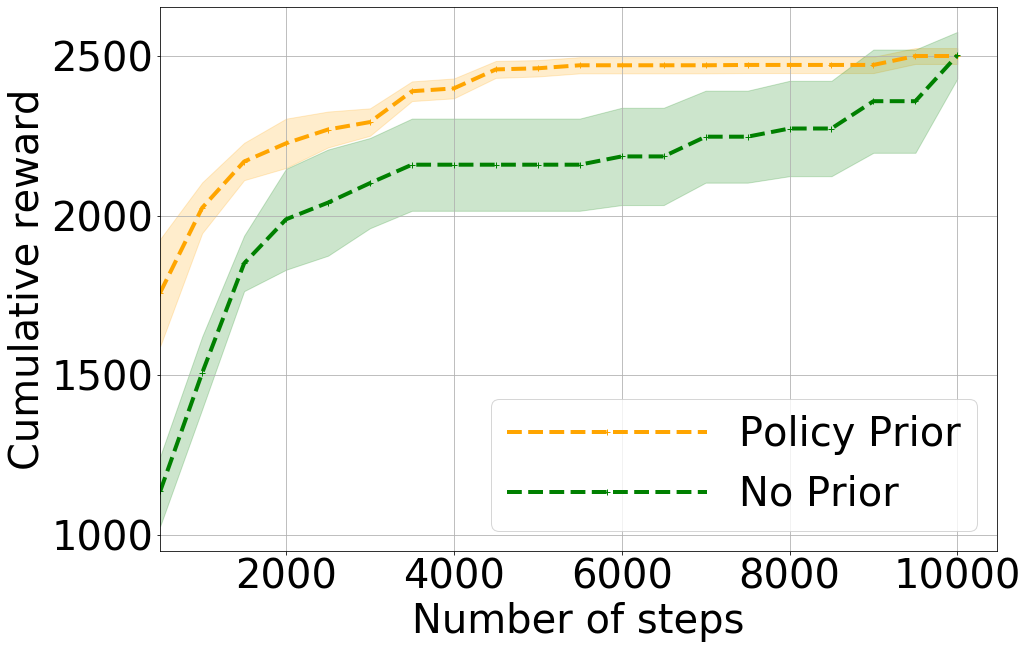}
}
\caption{\label{fig:gym-2d}\label{fig:gym-5d}Performance (cumulative reward) using a BO policy search with \emph{Policy prior} on three MuJoCo tasks a) when there are \textbf{two} (friction and mass), and c, d) \textbf{five} (friction, restitution and mass of 3 body parts) unknown latent parameters of Hopper, Walker2D and Half-Cheetah. For all tasks a 500 step horizon is used.}
\end{figure*}


Fig. \ref{fig:gym-2d} and Table \ref{table:results-mujoco} illustrate the sample
efficiency of the \emph{policy prior} introduced in Section \ref{subsec:Building-the-gp-prior}
when integrated with BO policy search, evaluated on the three MuJoCo environments. For each environment we consider two settings with two (2 Dim) and five (5 Dim) unknown physics latent factors are present in the real-world, which demonstrate the efficacy
of our method in both low and moderately high dimensional latent factor space. 

When building the priors for these environments, in the 2 Dim setting
we uniformly sample $100$ latent factors while in 5 Dim
setting $250$ samples are used. We then evaluate the performance
of UPN policies fetched at each sampled latent value on $10$ and $250$ different simulated
environments (i.e., environments with different true latent values) for 2 Dim and 5 Dim settings, respectively. 
For both Bowling and Basketball tasks, we follow the same procedure to build priors for the contextual bandit agent, using $100$ latent factor combinations drawn by discretizing the latent space uniformly.



\subsubsection*{Evaluation using low cost fidelities}

\begin{table*}
\begin{centering}
\subfloat[\label{table:results-mujoco} Jump start performance (cumulative episodic reward) with \emph{Policy prior} on
three MuJoCo tasks when there are two (2 Dim) and five (5 Dim) unknown
latent parameters. For all tasks a 500 step horizon is used and jump start
is averaged over 100 random episodes.]{\centering{}%
\begin{tabular}{|c|c|c|c|c|}
\hline 
 & MAML & DR & \emph{No Prior} & \emph{Policy Prior}\tabularnewline
\hline 
\hline 
Walker2D - 2 Dim & 514.70 $\pm$ 1.41 & 1530.18 $\pm$ 23.56 & 1353.82 $\pm$ 6.32 & \textbf{1602.9 $\pm$ 6.54}\tabularnewline
\hline 
Walker2D - 5 Dim & 552.24 $\pm$ 15.30 & 941.88 $\pm$ 18.17 & 1146.98 $\pm$ 6.00 & \textbf{1186.76 $\pm$ 6.53}\tabularnewline
\hline 
Half-Cheetah - 2 Dim & 592.76 $\pm$ 39.81 & \textbf{4581.58 $\pm$ 22.38} & 3986.11 $\pm$ 9.39 & 4050.92 $\pm$ 6.07\tabularnewline
\hline 
Half-Cheetah - 5 Dim & 375.54 $\pm$ 3.85 & 1784.36 $\pm$ 9.33 & 2261.26 $\pm$ 11.1 & \textbf{2310.88 $\pm$ 7.80}\tabularnewline
\hline 
Hopper - 2 Dim & 394.01 $\pm$ 1.23 & 1603.51 $\pm$ 10.32 & \textbf{1930.11 $\pm$ 6.32} & \textbf{1928.74 $\pm$ 5.46}\tabularnewline
\hline 
Hopper - 5 Dim & 658.21 $\pm$ 3.69 & 1322.75 $\pm$ 5.90 & 1519.73 $\pm$ 5.06 & \textbf{1541.45 $\pm$ 3.96}\tabularnewline
\hline 
\end{tabular}}

\subfloat[\label{table:results-bandits} Jump start performance (\emph{AUCCESS} score) of bandit tasks. Policy search process has been conducted using low fidelity actions, 20 for Bowling and 5 for basketball.]{\centering{}%
\begin{tabular}{|c|c|c|c|c|}
\hline 
 & DR & Estimated & \emph{No Prior} & \emph{Policy Prior}\tabularnewline
\hline 
\hline 
Bowling & 0.8953$\pm$ 0.0067 & 0.8956$\pm$ 0.0108 & 0.8985$\pm$ 0.0040 & \textbf{0.9202$\pm$ 0.0053}\tabularnewline
\hline 
Basketball & 0.7390 $\pm$ 0.0723 & 0.3145 $\pm$ 0.1147 & \textbf{0.8784 $\pm$ 0.0827} & \textbf{0.9519 $\pm$ 0.011}\tabularnewline
\hline 
\end{tabular}}

\caption{\label{table:results-all}\emph{Policy Prior's} initial transfer policy performance (jump start) on MuJoCo and Bandit tasks compared with baselines. Bold represents the best performance.}
\par\end{centering}
\end{table*}

One of the key issues with UPN based online system identification
is the evaluation cost during the policy search process. For instance, if
we follow the PHYRE formulation for Bowling task which evaluates 100 actions per task,
it would cost 40,000 interactions with the real-world to evaluate
20 BO iterations using a validation set of 20 tasks (100 actions $\times$
20 tasks $\times$ 20 iterations). To improve the sample efficiency
of this policy search process, we examine using a low fidelity evaluation by choosing the best 20 actions based on their associated rewards obtained from rollouts carried out in simulation. 
Following the policy search with the policy prior, we evaluate the performance of the latent conditioned policy using 100 best ranked actions on the test set (Table. \ref{table:results-bandits}). Similar evaluations are carried out for the Basketball task using 5 best actions for policy search and jump start evaluated using the best 100 actions. 




We designed the Basketball task such that coefficient of restitution (CoR) of objects are more impactful for the task's goal than friction. For example, when the CoR of the ball is approximately less than 0.3, the task cannot be achieved irrespective of the friction value. We found that the Gaussianity checking step of the \emph{Policy prior} (Sec. \ref{subsec:gaussianity-check}) is particularly useful to address such non-Gaussian settings (refer appendix for further details).

From the results in Fig. \ref{fig:gym-5d} and Table \ref{table:results-all}, it is clear that considering the superior sample efficiency offered through policy search, combined with the consistently good jump start performances, the prior policy is a beneficial tool for efficient learning in the parameterisable environments considered in this work. For low dimensional cases (e.g. 2 Dim), DR performs on par or better than BO methods, likely due to the low randomness offered by the environment. However, when the number of unknown parameters increases, DR performs poorer than policy search methods.

%% file: bg.tex
As a solution to sample inefficiency in many deep RL applications,
efficient simulation based transfer (i.e., Sim2Real) has become a
necessary, albeit a challenging task. The fundamental requirement
for such transfer is that simulation and real-world having some shared
basis for transfer, which has been studied in different granularities.
One such simulation-in-the-loop approach, as shown by \cite{using-inaccurate-models,farchy2013humanoid,10.5555/3304889.3305112,DBLP:conf/icra/ChebotarHMMIRF19,du2021auto, allevato2020tunenet,Allevato:2020ui,gat2017,stocgat20},
is system identification, which adapts the simulator to closely match
the real-world trajectories (i.e., `grounding',) by minimising the
trajectory difference (i.e., residual) between the model and real-world.
As an optimisation workflow, they iteratively execute a simulator
trained policy in the real-world and improve simulated latent factor
estimates. However, these attempts suffer from
several key issues: a) residual minimisation is unaware of goal-conditioned
latent factors, which may overexert the model learning process; b)
designing a residual measure that guarantees to converge to the true
latent estimate is challenging as shown by \cite{DBLP:conf/icra/ChebotarHMMIRF19},
who used a weighted importance measure for this purpose; c) iterative
policy training in simulator and real-world is not practical for time-sensitive
tasks; and d) latent estimation being sensitive to the initialisation,
can lead to suboptimal estimations.


To address some of the issues with system identification, task conditioned
policy learning methods have been proposed. One such approach is domain
randomisation (DR), which trains agents in a range of randomised latent
factors to find invariant task policies through different dynamics
\cite{8202133,DBLP:conf/rss/SadeghiL17,matas2018sim} . While it
is relatively simple to implement, this method has shown to produce
overly robust and conservative policies that may not be capable of
quickly adapting to a given environment \cite{data-driven-dr}. Moreover,
\cite{matas2018sim} observed that over randomisation can have an
detrimental effect on DR learning, which raises the difficulty of
identifying the boundaries of DR training. 

Instead of learning conservative policies, a line of research has
studied evaluating policies directly in the real-world task (i.e.,
direct policy search) to improve the simulated model \cite{evalutionary-direct-search}.
However, this introduced the difficulty of iteratively training simulated
policies and a likely high evaluation cost in the real-world. In a novel
approach, \cite{up-net} introduced Universal Policy networks (UPN),
an RL agent model trained to learn a large set of policies, each of
which is conditioned on a latent factor. They showed this model was
capable of generalising to estimate policies for unseen latent values.
To estimate latent factors from observations, they used a DL based
system identification model, but \cite{yu2019sim,yu2018policy} successfully
demonstrated using direct policy search with UPN by estimating latent
factors from an evaluated real-world policy such that the policy emitted
from UPN for estimated latent values will improve the policy transfer.
To estimate the latent values, they used Bayesian Optimisation (BO)\cite{frazier2018tutorial}
with Gaussian Processes (GP) \cite{Rasmussen2004-wq} as a cheap
surrogate function of the true function. Our approach, while using a similar
workflow, improves the sample efficiency of policy search process over these methods
by instilling prior information about the policy and dynamic function behaviours. 


Meta learning is another line of research that addresses fast adaptation
to a new task. Model Agnostic meta-learning (MAML)\cite{finn2017model} introduces a gradient based
parameters optimisation method for a set of tasks such that adapting to a new task is sample efficient.
Nagabandi et al.\cite{nagabandi2018learning} extends this concept to learn a model prior capable of rapidly adapting to a new task, whereas, 
Mendonca et al. \cite{mendonca2020meta} and Yu et al.\cite{yu2020learning} meta trains a context variable that can be
used as a latent factor to condition a trained UPN. We use MAML as one of our baselines to examine the sample efficiency
of meta-learning when adapting to a new environment.

%% file: conclude.tex
In this work, we proposed a novel model-based prior to improve the
sample efficiency of direct policy search when learning in an unknown
environment. Using a numerical simulator to evaluate
universal policy network (UPN) based policies in varied simulated environments, our proposed approach was used to obtain an estimate of its inter-policy similarity. 
This was then integrated with a Gaussian Process based
Bayesian Optimisation workflow in order to efficiently identify appropriate policies from the UPN. We empirically evaluated our method
in five MuJoCo and bandit learning environments, and demonstrated its
superior sample efficiency in terms of search convergence and transfer
jump start performances compared to competing baselines.


%% file: supp.tex
\subsection{Policy Prior Integrated Policy Search Performances - Extended}

\begin{figure*}
\centering{}
\subfloat[2Dim-Walker2D]{\includegraphics[width=0.30\paperwidth]{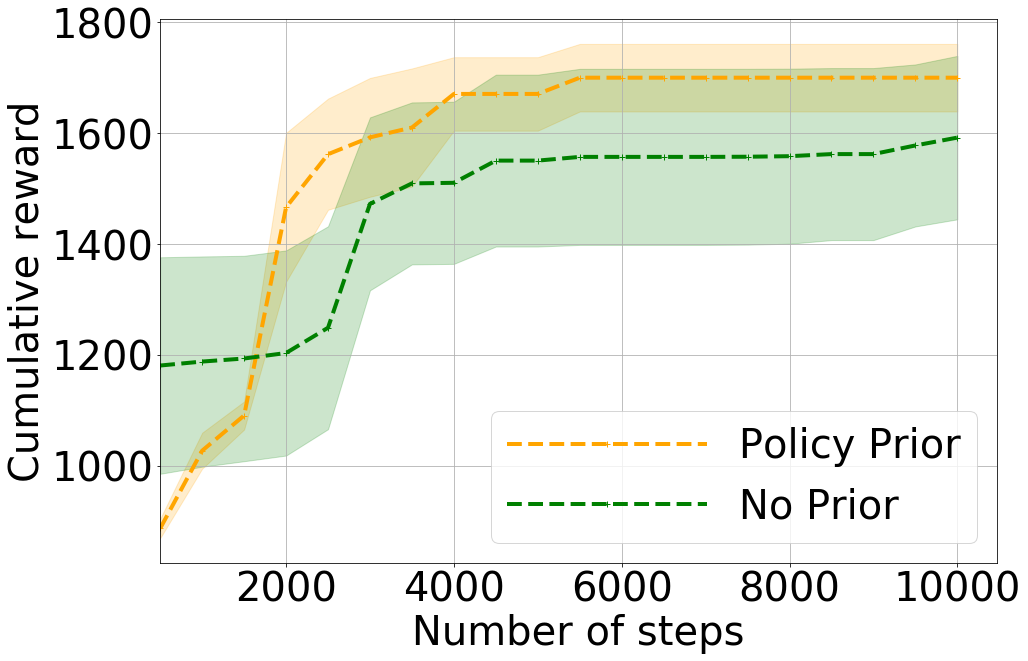}
}
\subfloat[5Dim-Walker2D]{\includegraphics[width=0.30\paperwidth]{imgs/5D/walker-5d-500win.png}
}
\medskip{}
\subfloat[2Dim-Half-Cheetah]{\includegraphics[width=0.30\paperwidth]{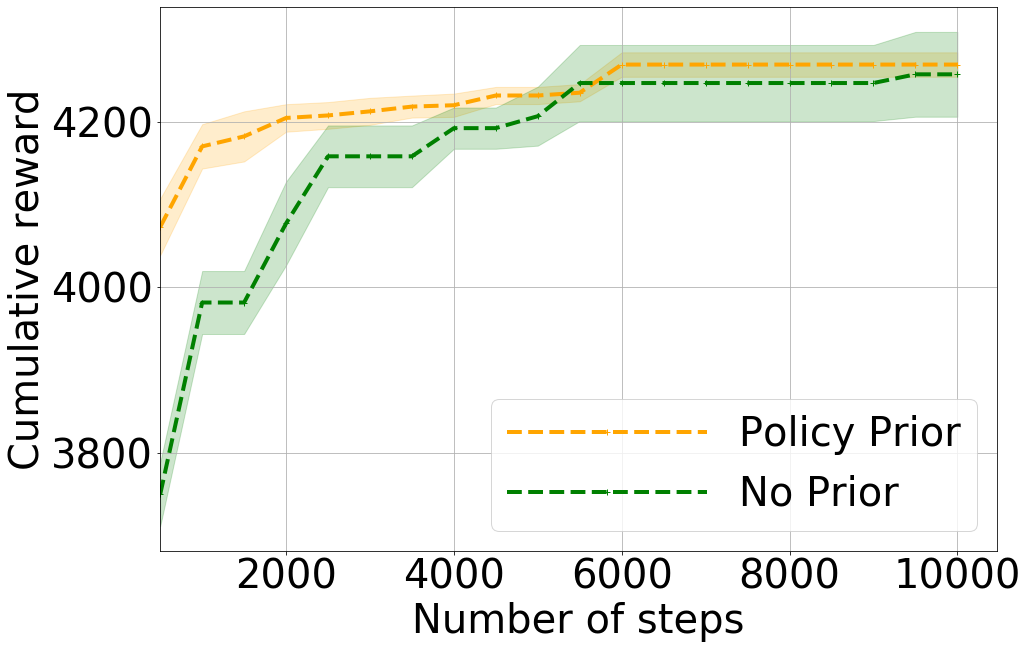}
}
\subfloat[5Dim-Half-Cheetah]{\includegraphics[width=0.30\paperwidth]{imgs/5D/cheetah-5d-500win}
}

\medskip{}

\subfloat[2Dim-Hopper]{\includegraphics[width=0.30\paperwidth]{imgs/2D/hopper-2d-500win.png}
}
\subfloat[5Dim-Hopper]{\includegraphics[width=0.30\paperwidth]{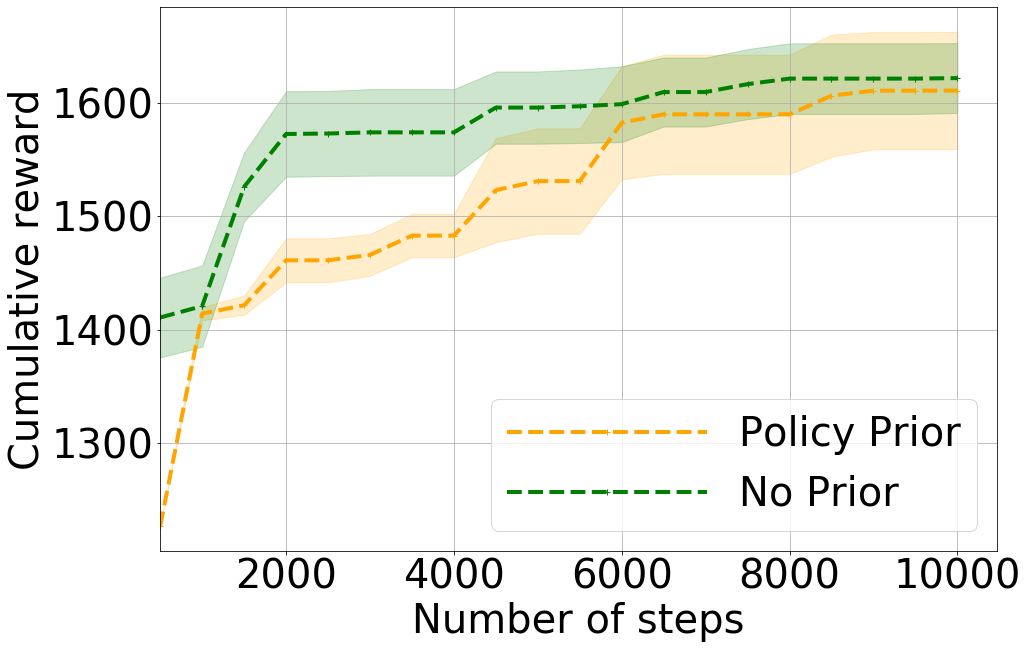}
}

\caption{\label{fig:gym-2d} Performance in terms of the cumulative reward using a Bayesian Optimisation policy search with \emph{Policy prior} on three MuJoCo tasks, a,c,e) when there are \textbf{two} unknown latent parameters,
mass and friction in the real world, and b,d,f) when there are \textbf{five} unknown latent parameters: friction, restitution and mass of 3 body parts
Hopper, Walker2D and Half-Cheetah. For all tasks a 500 step horizon is used.}
\end{figure*}

\begin{figure*}
\begin{centering}
\subfloat[Bowling\label{fig:phyre-bo-action5}]{\includegraphics[width=0.30\paperwidth]{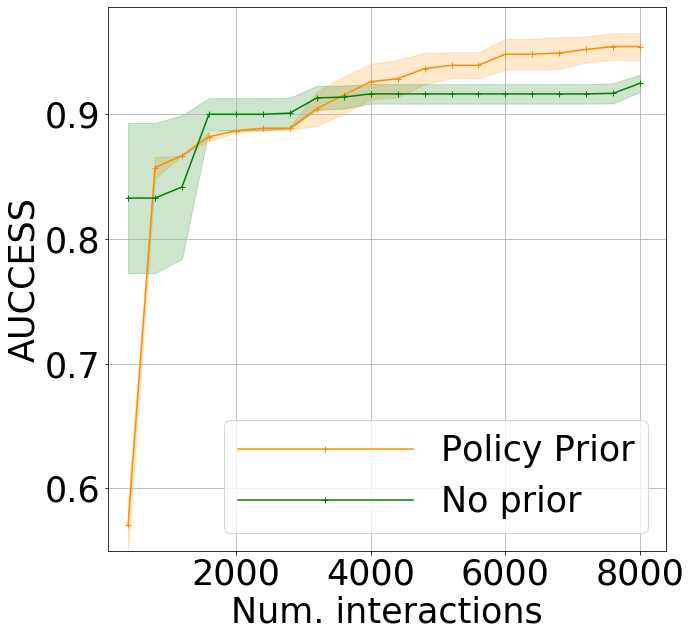}
}\qquad{}
\subfloat[Basketball\label{fig:gaussian-check-bb}]{\includegraphics[width=0.30\paperwidth]{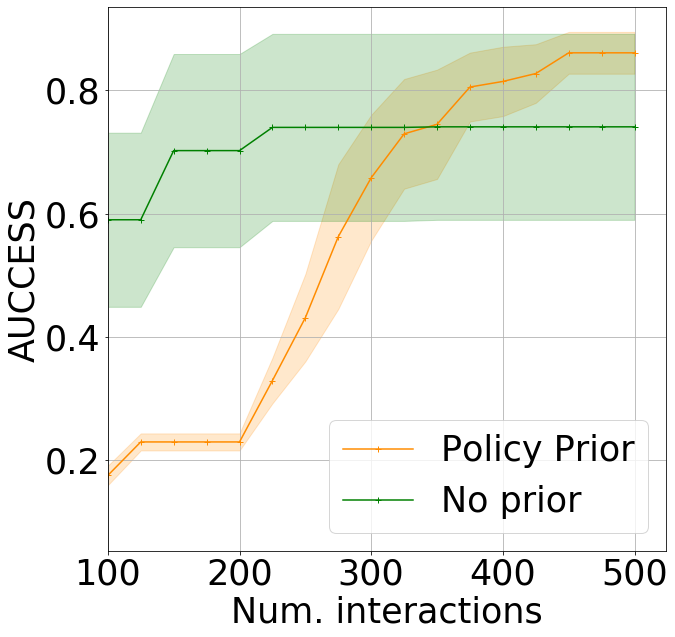}
}\qquad{}
\par\end{centering}
\caption{\label{fig:low-fidelity}\emph{Policy Prior's} policy search performances in terms of the \emph{AUCCESS} score, on the validation task set. Policy search has been conducted using a low fidelity action set for bandit tasks, (a) Bowling task with 20 best actions and (b) Basketball task with 5 best actions.
}
\end{figure*}

Fig. \ref{fig:gym-2d} shows the performance of \emph{Policy prior} integrated Bayesian Optimisation policy search, in terms of the cumulative reward, on three MuJoCo tasks, Hopper, Walker2D and Half-Cheetah. Fig. \ref{fig:low-fidelity} shows the \emph{Policy prior}'s policy search performance on two bandit tasks, Bowling and Basketball, in terms of \emph{AUCCESS} score \cite[Sec. 3.2]{Bakhtin2019-dq}.

\subsection{Ablation Study: Checking for Gaussianity of Prior in Basketball Task}

\begin{figure}
\begin{centering}
\subfloat[\label{fig:mean-basketball-9}]{\includegraphics[width=0.2\paperwidth]{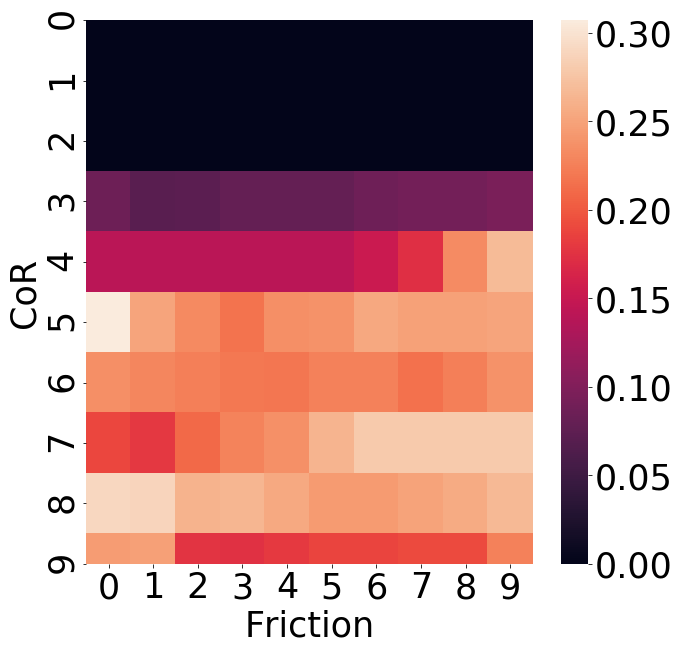}

}\subfloat[\label{fig:variance-basketball-9}]{\includegraphics[width=0.2\paperwidth]{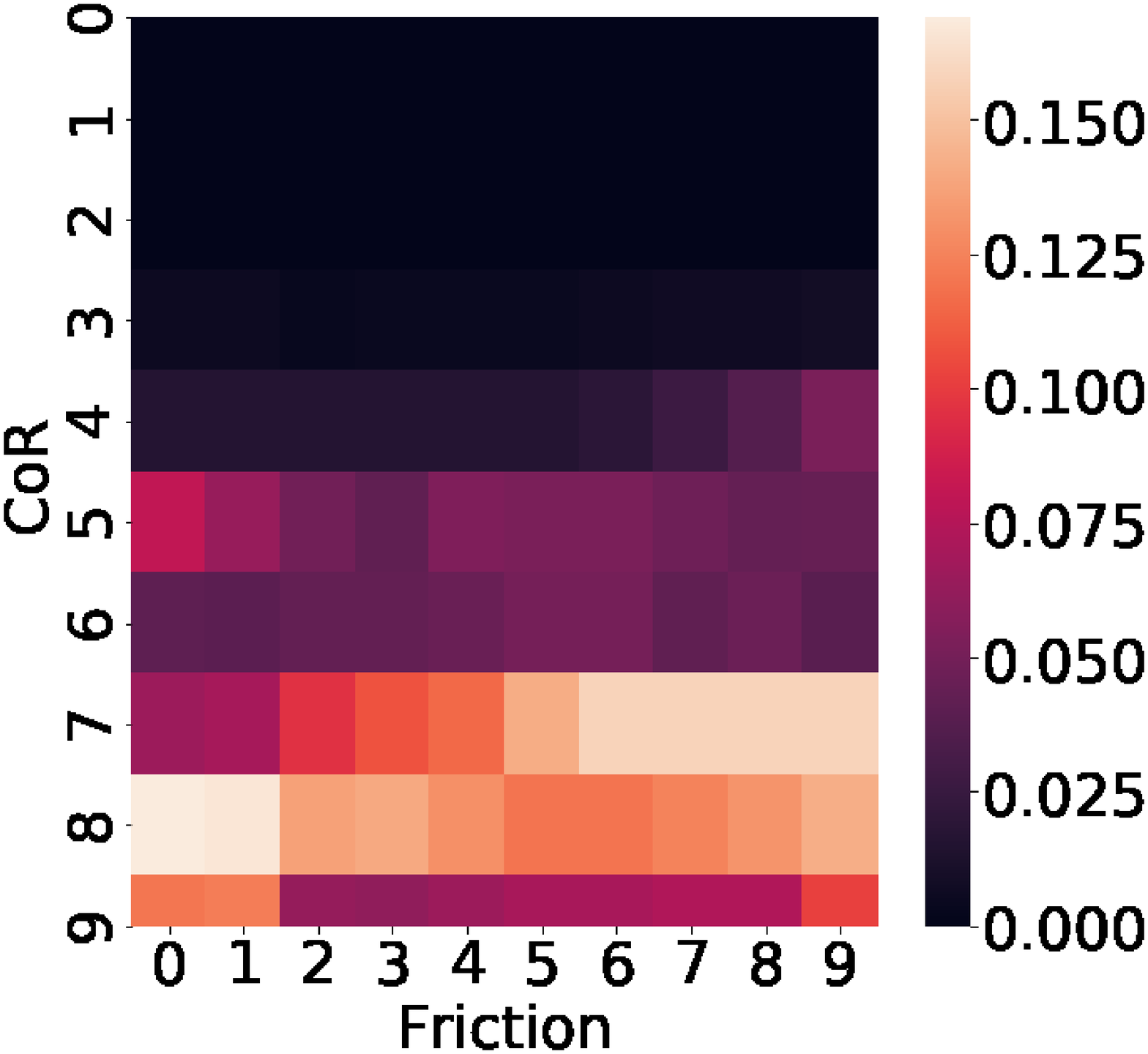}
}\medskip{}
\subfloat[\label{fig:bo-converg-bb-no-filter}]{\includegraphics[width=0.25\paperwidth]{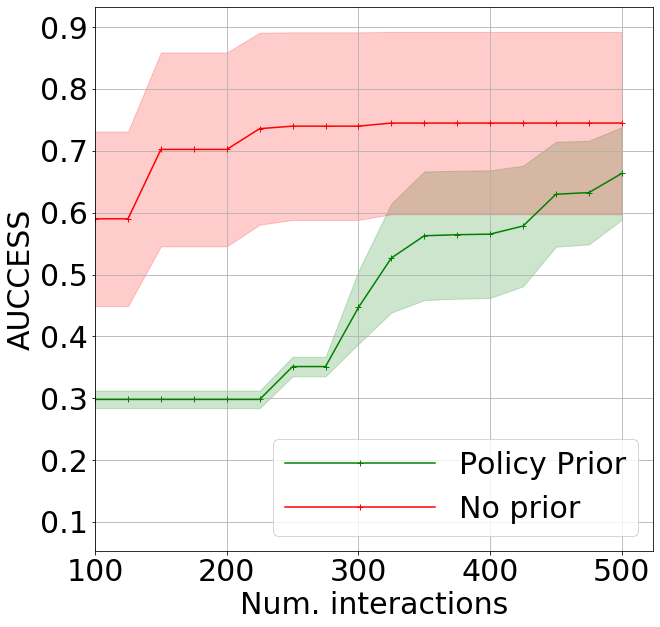}

}
\par\end{centering}
\caption{\label{fig:non-gaussian-task}Basketball task's non-Gaussian latent
factor distribution's effect on BO. a) Mean and b) variance of
jump start performances of basketball UPN policies when conditioned
at 10$\times$10 grid of latent values and evaluated across 9 different
ground truth settings (combinations of friction and CoR at 0.2, 0.5,
0.8). c) \emph{Policy prior's} BO convergence performance in Basketball
task without the Gaussianity check, 
using only the 5 best actions at each evaluation. All evaluations
are executed in the simulated environment with Basketball validation
test set.}
\end{figure}

Basketball task is designed such that CoR of objects are more impactful
for the task's goal than friction. This is elucidated in Fig. \ref{fig:mean-basketball-9}
and \ref{fig:variance-basketball-9}, which show the mean and variance
of jump start performances of basketball UPN policies, when conditioned
at 10$\times$10 grid of latent values and then transferred to 9 different
ground truth settings. In essence, when the CoR of the ball is approximately
less than 0.3, the task can not be achieved irrespective of the friction
value. We utilise this skewed setting to
investigate the behaviour of our \emph{Policy
prior} in tasks that do not respond equally to all latent factors
involved. 

Fig. \ref{fig:bo-converg-bb-no-filter} shows that during BO policy search, 
\emph{Policy Prior} without the Gaussianity check performing 
worse than the no prior baseline on the Basketball task. We posit this is due to the non-Gaussian
distribution of the \emph{Policy Prior} and to remedy this issue, we filter out non-Gaussian
observations with p-value less than 0.1 using Kolmogorov-Smirnov Goodness-of-Fit
Test. In the best 5 action setting of the basketball task we are using,
it filters out 59 observations out of 100. With this modified prior,
we run the trial again and Fig. \ref{fig:gaussian-check-bb} shows
the improved BO policy search performance. This relatively simple step can assist
in making the \emph{Policy Prior} useful for tasks irrespective
of their latent factor distribution. However, it comes with a caveat
that the loss of observations could reduce the effect of the prior
if most of the observations are filtered out.

\subsection{\label{sec:dqn_upn}DQN Based Universal Policy Network}

For training UPNs for contextual bandit tasks, we use Deep Q-Networks \cite{mnih2015human} as the
underlying agent. When implementing the DQN based UPN, we augment a DQN architecture
adopted from PHYRE \cite{Bakhtin2019-dq} by supporting conditioning
with latent factors in addition to actions (Fig. \ref{fig:dqn-upn}).
For this purpose we use a FiLM layer \cite{perez2018film} due to
its ability to modulate between two signals. We train this model on
simulator generated trajectories by sampling latent factors for a
given task. 

\begin{center}
\begin{figure}
\begin{centering}
\includegraphics[width=0.3\paperwidth]{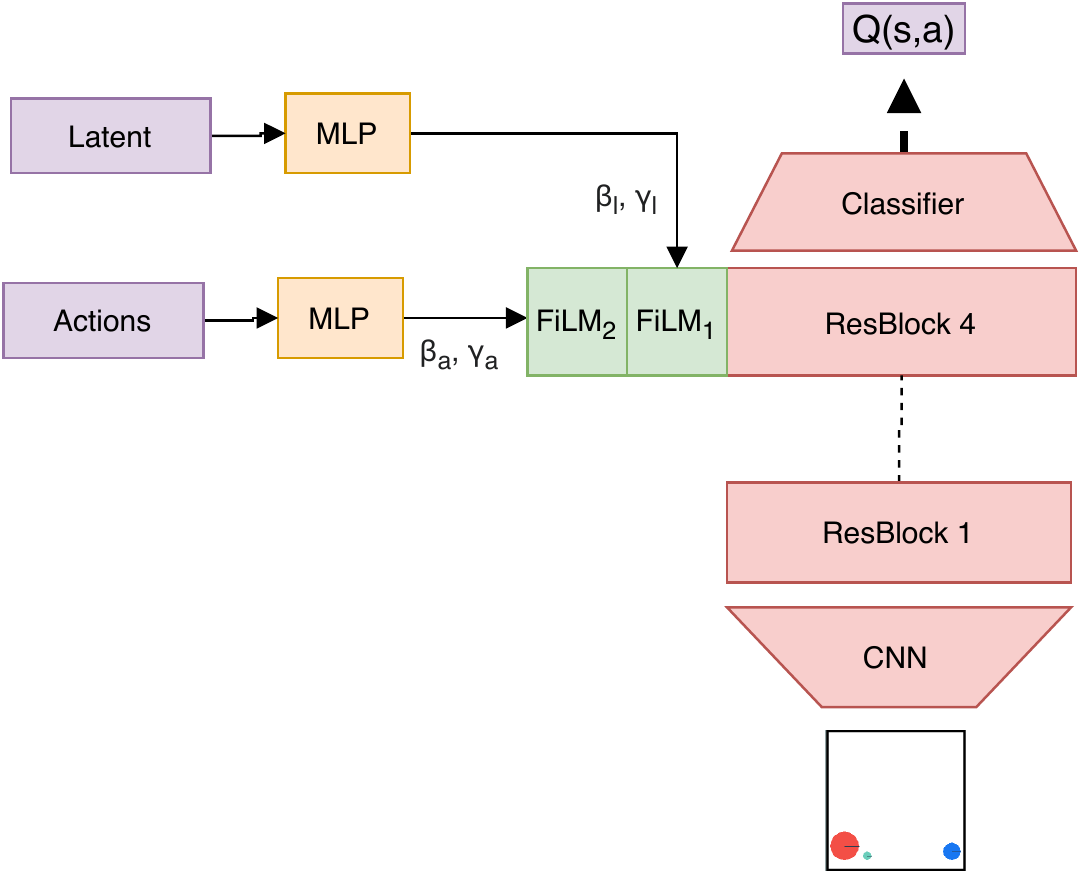}
\par\end{centering}
\caption{\label{fig:dqn-upn} A DQN based UPN design. Latent and action vectors
are encoded using MLP layers, which are used to condition the observations
encoded by a residual network \cite{he2016deep} using FiLM layer
\cite{perez2018film}. $\beta$ and $\gamma$ are parameters learnt
by FiLM to modulate conditioning.}
\end{figure}
\par\end{center}

\subsection{Universal Policy Network Training}

We use the DQN based UPN implementation introduced in Sec. \ref{sec:dqn_upn}
to find policies for our contextual bandit tasks. In these tasks,
we keep the density of objects consistent and maintain friction and
coefficient of restitution (CoR) as the unknown latent factors. When
training the UPN, we select 12 combinations of friction and CoR in
the range of {[}0, 1{]} from a grid with resolution of 0.3 as conditioning
latent points. We then train the UPN on tasks in the training fold
of dataset, each of which is replicated on all 12 latent factor combinations 
(i.e., each task is simulated under all sampled latent factor settings).
In order to decide when the UPN has sufficiently converged to an overall
good policy, we evaluate the UPN's policy at 100 sampled latent value
points, which are then transferred to 9 different simulated ground
truth latent values as shown in Fig. \ref{fig:UPN-convergence}. A
sufficiently converged UPN policy should give relatively high rewards
when the ground truth latent values approximately overlap with the
sampled latent values in the UPN. Leveraging this strategy, we selected
UPN models trained for 32 batches of 24,000 and 48,000 updates for
Bowling and Basketball tasks respectively. Furthermore, this verification
establishes that our DQN based UPN can learn latent conditioned policies
successfully.
\begin{center}
\begin{figure}
\begin{centering}
\includegraphics[width=0.4\paperwidth]{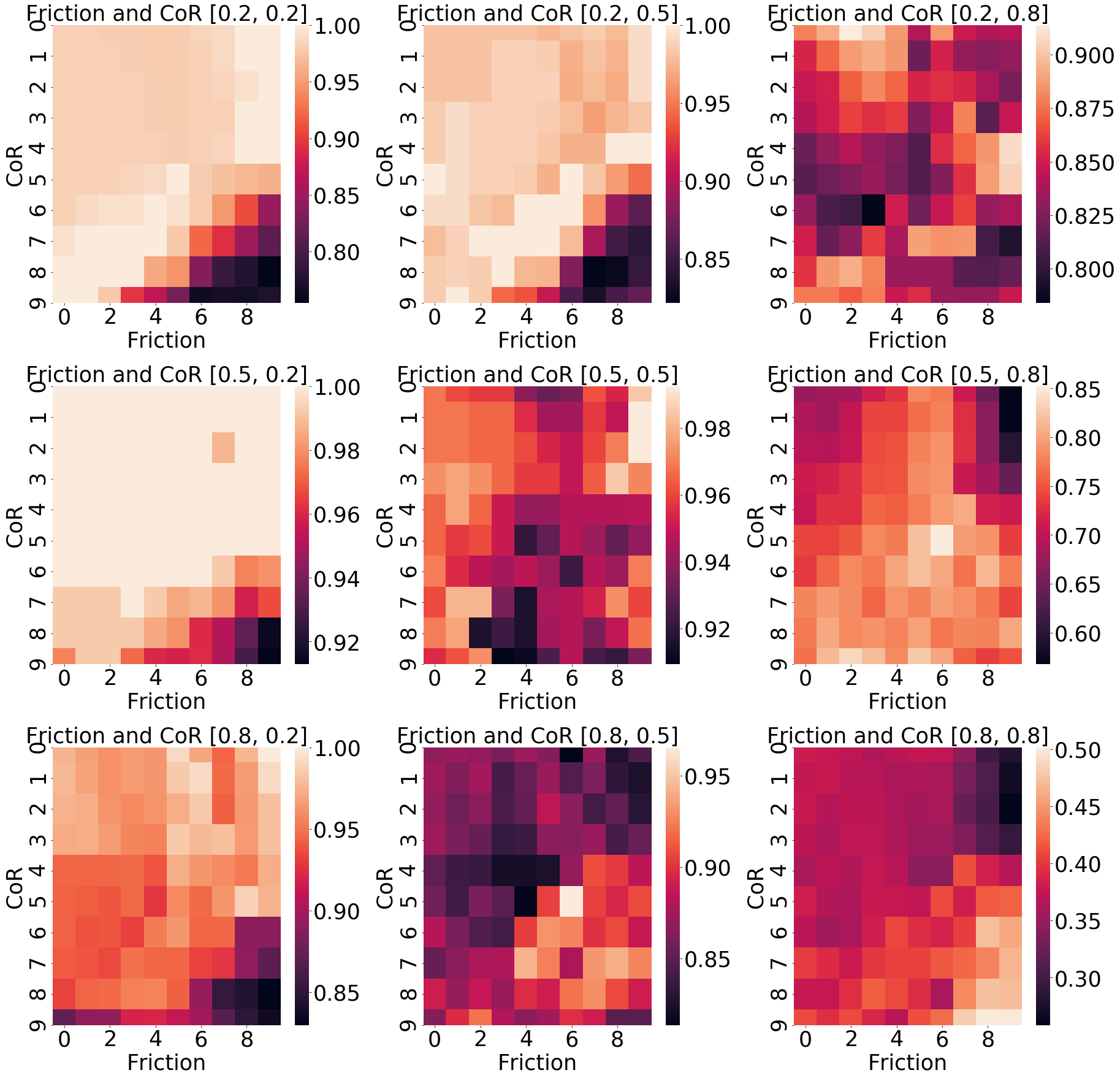}
\par\end{centering}
\caption{\label{fig:UPN-convergence}A simple strategy to decide whether the
UPN has converged. 100 sampled latent value combinations of friction
and CoR are evaluated on the UPN to select a policy and then transferred
to 9 different environment settings. The title of each subplot refers
to the ground truth friction and CoR value of the environment in which
the policy is being evaluated at. A well trained UPN should provide
policies that yield high rewards when UPN latent values and ground
truth values match.}
\end{figure}
\par\end{center}

\subsection{UPN Training performances}

\begin{figure*}
\centering{}
\subfloat[Hopper-2D]{\includegraphics[width=0.30\paperwidth]{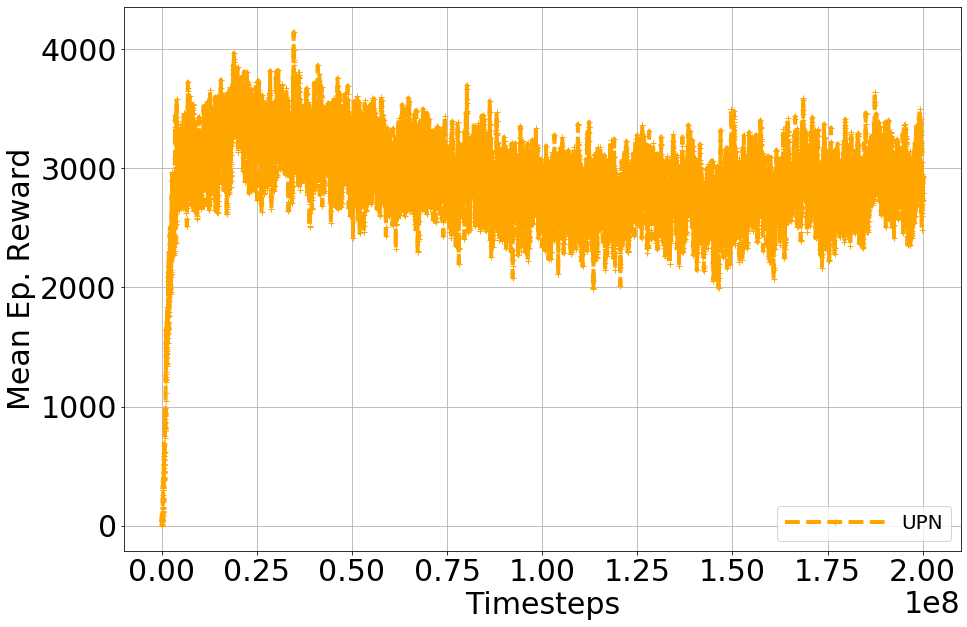}

}\subfloat[Hopper-5D]{\includegraphics[width=0.30\paperwidth]{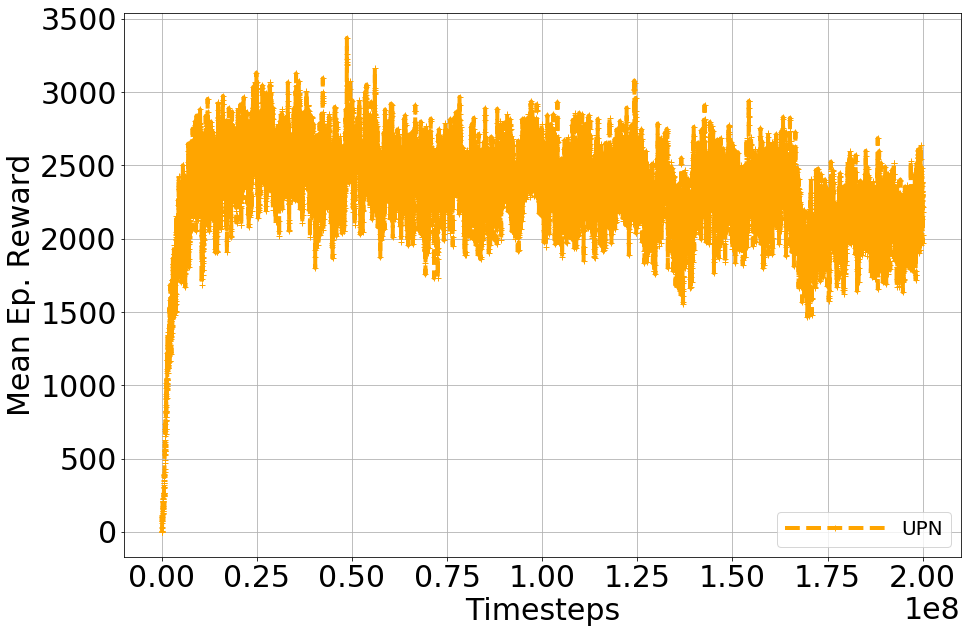}

}\medskip{}
\subfloat[Walker-2D]{\includegraphics[width=0.30\paperwidth]{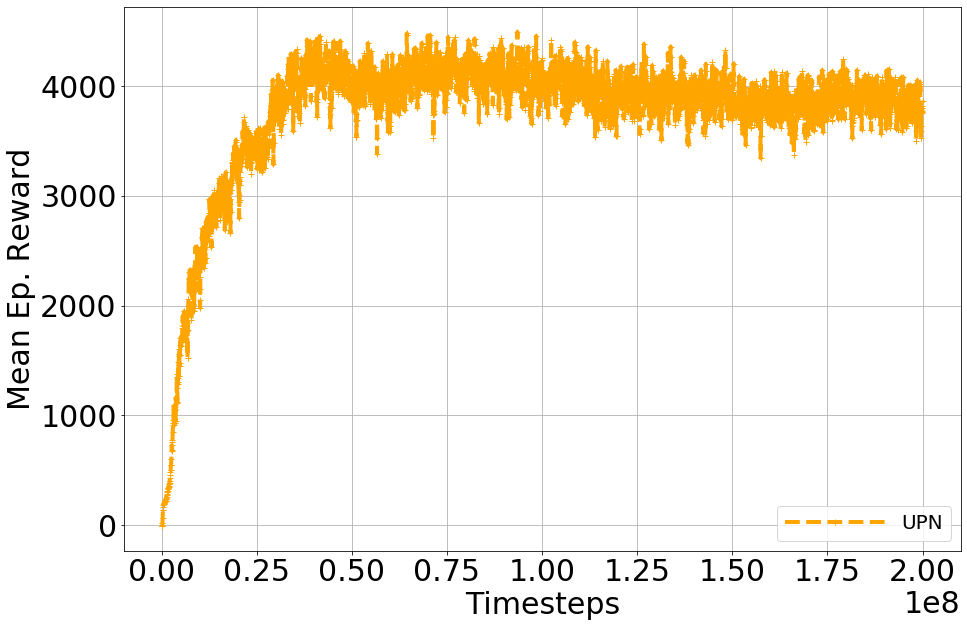}

}\subfloat[Walker-5D]{\includegraphics[width=0.30\paperwidth]{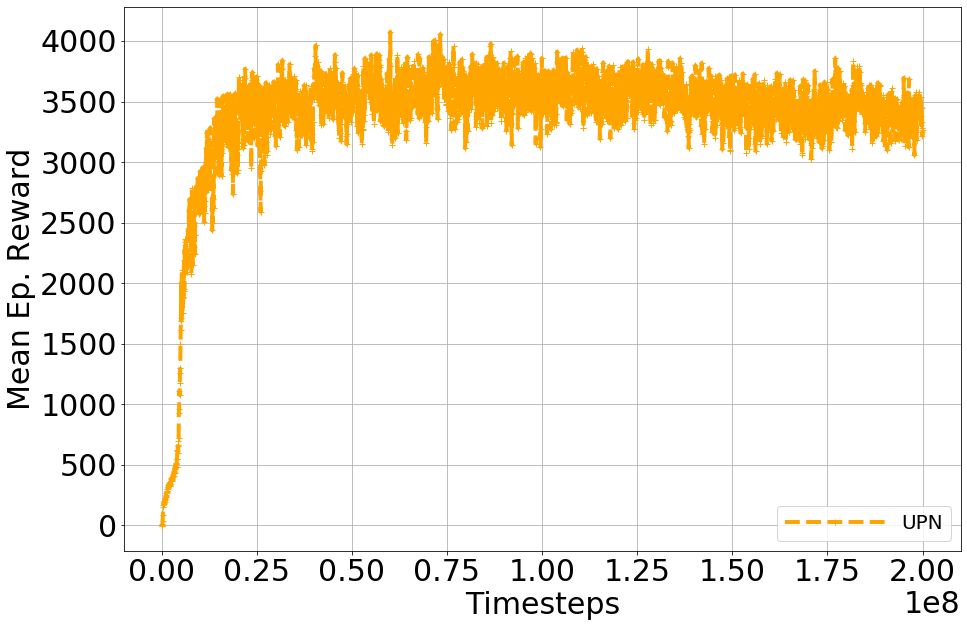}

}\medskip{}
\subfloat[Half-Cheetah-2D]{\includegraphics[width=0.30\paperwidth]{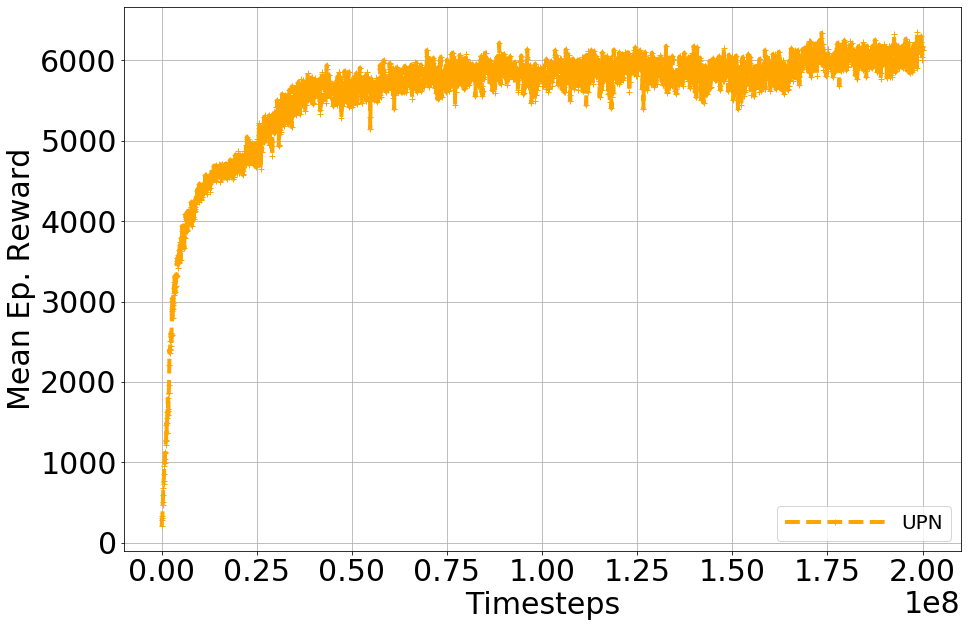}
}
\subfloat[Half-Cheetah-5D]{\includegraphics[width=0.30\paperwidth]{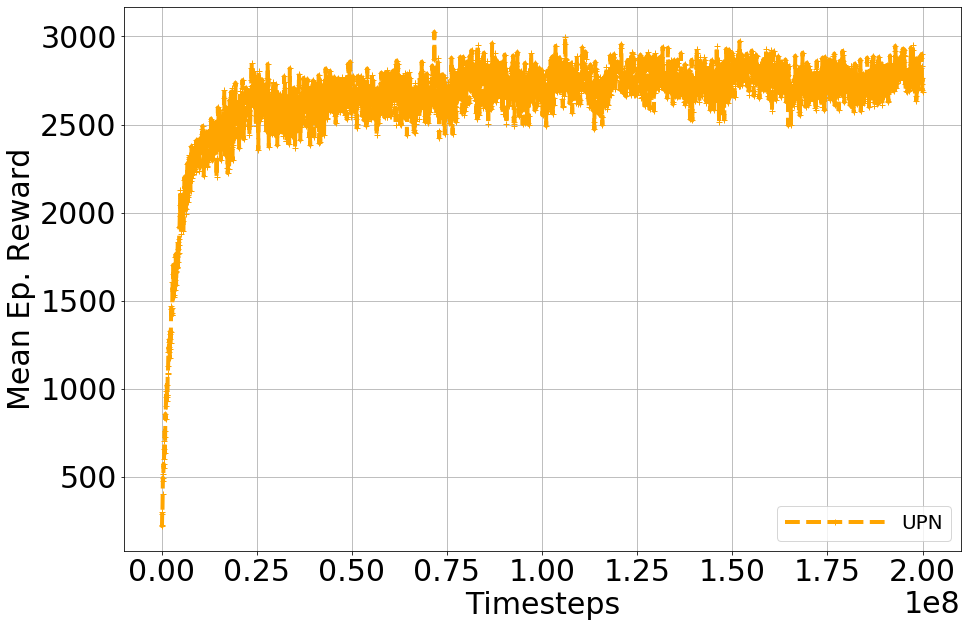}
}
\caption{\label{fig:upn-training} UPN average episodic reward during training (in simulation) for three MuJoCo tasks: Hopper, Walker2D and Half-Cheetah.  a, c, e) 2 unknown latent factors (2D) setting (mass and friction), and b, d, f) 5 unknown latent factors (5D) setting (friction, restitution and mass of 3 body parts).}
\end{figure*}

Fig. \ref{fig:upn-training} shows the training performances of Universal Policy Networks (UPN) on three MuJoCo environments,
each under two settings of two (2D) and five (5D) unknown latent factors. In the 2D setting, mass of a body node and friction of the 
tasks are unknown, whereas in the 5D setting, friction, restitution and mass of 3 body parts are unknown.

\subsection{Implementation Details}

When conducting BO based policy search, we used GPy\cite{gpy2014} to build the GP model 
and Emukit \cite{emukit2019} for Bayesian Optimisation execution.
All \emph{Policy Prior} and \emph{No Prior} results reported are averaged over 5 trials, each
of which is initialised with its random
number generator seed mutually exclusively set to one of 50, 100,
150, 500 and 1000. For all BO operations, we use RBF kernel, Expected
improvement (EI) as the acquisition function, with LBFGS as the optimiser. During BO, to maintain the matrix is well-conditioned, we verify the condition number of the kernel matrix is at most 25,000 and add random noise from a range of {[}0.1, $10^{-5}${]} to the diagonal if it exceeds this threshold.

For no prior baseline, the process starts
with 3 random samples and proceeds to build the GP. In MuJoCo tasks,
BO samples within the bounds {[}0, 1{]}, as all features are normalised.
For bandit tasks, friction range is {[}0, 3{]} while CoR is {[}0,
1{]}.

For MuJoCo tasks, depending on the setting (2D or 5D), we set the ground truth to a vector of length 2 or 5, 
generated with \emph{NumPy} random.uniform()\footnote{https://numpy.org/doc/stable/reference/random/generated/numpy.random.uniform.html} with random number generator seed set to 1. To separate the real-world from the simulated environment, 
we use a damping value of 3 using \emph{PyDart} simulator's \emph{set\_damping\_coefficient} API call. 

Contextual bandit tasks maintain standard three folds of task sets
and a maximum action size to sample from. The Bowling task uses 5000
actions, with 100 tasks in total split to 60 training, 20 validation
and 20 test tasks. The Basketball task utilises
40,000 actions, with a dataset containing 25 tasks, split into 15
training, 5 validation and 5 test sets. When setting up the real-world,
for both tasks a damping of 0.8 is used whereas in the simulated environment
no damping value is used. For the \emph{Estimated} baseline,  6,000 real-world interactions for the Bowling task and 1,500 for the Basketball task have been used to minimise the difference between real-world and simulated trajectories.

%% file: main.bbl
\begin{thebibliography}{10}
\providecommand{\url}[1]{#1}
\csname url@samestyle\endcsname
\providecommand{\newblock}{\relax}
\providecommand{\bibinfo}[2]{#2}
\providecommand{\BIBentrySTDinterwordspacing}{\spaceskip=0pt\relax}
\providecommand{\BIBentryALTinterwordstretchfactor}{4}
\providecommand{\BIBentryALTinterwordspacing}{\spaceskip=\fontdimen2\font plus
\BIBentryALTinterwordstretchfactor\fontdimen3\font minus
  \fontdimen4\font\relax}
\providecommand{\BIBforeignlanguage}[2]{{%
\expandafter\ifx\csname l@#1\endcsname\relax
\typeout{** WARNING: IEEEtran.bst: No hyphenation pattern has been}%
\typeout{** loaded for the language `#1'. Using the pattern for}%
\typeout{** the default language instead.}%
\else
\language=\csname l@#1\endcsname
\fi
#2}}
\providecommand{\BIBdecl}{\relax}
\BIBdecl

\bibitem{silver2016mastering}
D.~Silver, A.~Huang, C.~J. Maddison, A.~Guez, L.~Sifre, G.~Van Den~Driessche,
  J.~Schrittwieser, I.~Antonoglou, V.~Panneershelvam, M.~Lanctot \emph{et~al.},
  ``Mastering the game of go with deep neural networks and tree search,''
  \emph{nature}, vol. 529, no. 7587, pp. 484--489, 2016.

\bibitem{mnih2015human}
V.~Mnih, K.~Kavukcuoglu, D.~Silver, A.~A. Rusu, J.~Veness, M.~G. Bellemare,
  A.~Graves, M.~Riedmiller, A.~K. Fidjeland, G.~Ostrovski \emph{et~al.},
  ``Human-level control through deep reinforcement learning,'' \emph{nature},
  vol. 518, no. 7540, pp. 529--533, 2015.

\bibitem{using-inaccurate-models}
\BIBentryALTinterwordspacing
P.~Abbeel, M.~Quigley, and A.~Y. Ng, ``Using inaccurate models in reinforcement
  learning,'' in \emph{Proceedings of the 23rd International Conference on
  Machine Learning}, ser. ICML '06.\hskip 1em plus 0.5em minus 0.4em\relax New
  York, NY, USA: Association for Computing Machinery, 2006, pp. 1--8. [Online].
  Available: \url{https://doi.org/10.1145/1143844.1143845}
\BIBentrySTDinterwordspacing

\bibitem{farchy2013humanoid}
A.~Farchy, S.~Barrett, P.~MacAlpine, and P.~Stone, ``Humanoid robots learning
  to walk faster: From the real world to simulation and back,'' in
  \emph{AAMAS}, 2013, pp. 39--46.

\bibitem{Zheng2018-bk}
D.~Zheng, V.~Luo, J.~Wu, and J.~B. Tenenbaum, ``Unsupervised learning of latent
  physical properties using {Perception-Prediction} networks,'' in
  \emph{Proceedings of the {Thirty-Fourth} {UAI} 2018, Monterey, California,
  {USA}, August 6-10, 2018}, A.~Globerson and R.~Silva, Eds.\hskip 1em plus
  0.5em minus 0.4em\relax AUAI Press, 2018, pp. 497--507.

\bibitem{Chang2016-pp}
M.~B. Chang, T.~Ullman, A.~Torralba, and J.~B. Tenenbaum, ``A compositional
  {Object-Based} approach to learning physical dynamics,'' Dec. 2016.

\bibitem{DBLP:conf/rss/Xu0ZTS19}
\BIBentryALTinterwordspacing
Z.~Xu, J.~Wu, A.~Zeng, J.~B. Tenenbaum, and S.~Song, ``Densephysnet: Learning
  dense physical object representations via multi-step dynamic interactions,''
  in \emph{Robotics: Science and Systems XV, University of Freiburg, Freiburg
  im Breisgau, Germany, June 22-26, 2019}, A.~Bicchi, H.~Kress{-}Gazit, and
  S.~Hutchinson, Eds., 2019. [Online]. Available:
  \url{https://doi.org/10.15607/RSS.2019.XV.046}
\BIBentrySTDinterwordspacing

\bibitem{8202133}
J.~{Tobin}, R.~{Fong}, A.~{Ray}, J.~{Schneider}, W.~{Zaremba}, and P.~{Abbeel},
  ``Domain randomization for transferring deep neural networks from simulation
  to the real world,'' in \emph{2017 IEEE/RSJ International Conference on
  Intelligent Robots and Systems (IROS)}, 2017, pp. 23--30.

\bibitem{DBLP:conf/rss/SadeghiL17}
F.~Sadeghi and S.~Levine, ``{CAD2RL:} real single-image flight without a single
  real image,'' in \emph{Robotics: Science and Systems XIII, MIT, USA, 2017},
  N.~M. Amato, S.~S. Srinivasa, N.~Ayanian, and S.~Kuindersma, Eds., 2017.

\bibitem{data-driven-dr}
M.~Sheckells, G.~Garimella, S.~Mishra, and M.~Kobilarov, ``Using data-driven
  domain randomization to transfer robust control policies to mobile robots,''
  in \emph{2019 International Conference on Robotics and Automation (ICRA)},
  2019, pp. 3224--3230.

\bibitem{yu2019sim}
W.~Yu, V.~C. Kumar, G.~Turk, and C.~K. Liu, ``Sim-to-real transfer for biped
  locomotion,'' in \emph{Proc. of The International Conference on Intelligent
  Robots and Systems (IROS)}, 2019.

\bibitem{lazaric2012transfer}
A.~Lazaric, ``Transfer in reinforcement learning: a framework and a survey,''
  in \emph{Reinforcement Learning}.\hskip 1em plus 0.5em minus 0.4em\relax
  Springer, 2012, pp. 143--173.

\bibitem{frazier2018tutorial}
P.~I. Frazier, ``A tutorial on bayesian optimization,'' \emph{arXiv preprint
  arXiv:1807.02811}, 2018.

\bibitem{joy2019flexible}
T.~T. Joy, S.~Rana, S.~Gupta, and S.~Venkatesh, ``A flexible transfer learning
  framework for bayesian optimization with convergence guarantee,''
  \emph{Expert Systems with Applications}, vol. 115, pp. 656--672, 2019.

\bibitem{openai-gym}
\BIBentryALTinterwordspacing
G.~Brockman, V.~Cheung, L.~Pettersson, J.~Schneider, J.~Schulman, J.~Tang, and
  W.~Zaremba, ``Openai gym,'' \emph{CoRR}, vol. abs/1606.01540, 2016. [Online].
  Available: \url{http://arxiv.org/abs/1606.01540}
\BIBentrySTDinterwordspacing

\bibitem{Bakhtin2019-dq}
A.~Bakhtin, L.~van~der Maaten, J.~Johnson, L.~Gustafson, and R.~Girshick,
  ``{PHYRE}: A new benchmark for physical reasoning,'' in \emph{NeurIPS}, 2019,
  pp. 5082--5093.

\bibitem{up-net}
\BIBentryALTinterwordspacing
W.~Yu, J.~Tan, C.~K. Liu, and G.~Turk, ``Preparing for the unknown: Learning a
  universal policy with online system identification,'' in \emph{Robotics:
  Science and Systems XIII, Massachusetts Institute of Technology, Cambridge,
  Massachusetts, USA, July 12-16, 2017}, N.~M. Amato, S.~S. Srinivasa,
  N.~Ayanian, and S.~Kuindersma, Eds., 2017. [Online]. Available:
  \url{http://www.roboticsproceedings.org/rss13/p48.html}
\BIBentrySTDinterwordspacing

\bibitem{yu2018policy}
\BIBentryALTinterwordspacing
W.~Yu, C.~K. Liu, and G.~Turk, ``Policy transfer with strategy optimization,''
  in \emph{International Conference on Learning Representations}, 2019.
  [Online]. Available: \url{https://openreview.net/forum?id=H1g6osRcFQ}
\BIBentrySTDinterwordspacing

\bibitem{ktest}
\BIBentryALTinterwordspacing
T.~S. community. (2008) scipy.stats.kstest. [Online]. Available:
  \url{https://docs.scipy.org/doc/scipy/reference/generated/scipy.stats.kstest.html}
\BIBentrySTDinterwordspacing

\bibitem{NIST-handbook}
N.~Heckert and J.~Filliben, ``\BIBforeignlanguage{en}{Nist/sematech e-handbook
  of statistical methods; chapter 1: Exploratory data analysis},'' 2003-06-01
  2003.

\bibitem{todorov2012mujoco}
E.~Todorov, T.~Erez, and Y.~Tassa, ``Mujoco: A physics engine for model-based
  control,'' in \emph{2012 IEEE/RSJ International Conference on Intelligent
  Robots and Systems}.\hskip 1em plus 0.5em minus 0.4em\relax IEEE, 2012, pp.
  5026--5033.

\bibitem{pymunk}
V.~Blomqvist, ``Pymunk simulator,'' http://www.pymunk.org/.

\bibitem{finn2017model}
C.~Finn, P.~Abbeel, and S.~Levine, ``Model-agnostic meta-learning for fast
  adaptation of deep networks,'' in \emph{International Conference on Machine
  Learning}.\hskip 1em plus 0.5em minus 0.4em\relax PMLR, 2017, pp. 1126--1135.

\bibitem{DBLP:journals/corr/SchulmanWDRK17}
\BIBentryALTinterwordspacing
J.~Schulman, F.~Wolski, P.~Dhariwal, A.~Radford, and O.~Klimov, ``Proximal
  policy optimization algorithms,'' \emph{CoRR}, vol. abs/1707.06347, 2017.
  [Online]. Available: \url{http://arxiv.org/abs/1707.06347}
\BIBentrySTDinterwordspacing

\bibitem{10.5555/3304889.3305112}
S.~Zhu, A.~Kimmel, K.~E. Bekris, and A.~Boularias, ``Fast model identification
  via physics engines for data-efficient policy search,'' in \emph{Proceedings
  of the 27th International Joint Conference on Artificial Intelligence}, ser.
  IJCAI'18.\hskip 1em plus 0.5em minus 0.4em\relax AAAI Press, 2018, pp.
  3249--3256.

\bibitem{DBLP:conf/icra/ChebotarHMMIRF19}
\BIBentryALTinterwordspacing
Y.~Chebotar, A.~Handa, V.~Makoviychuk, M.~Macklin, J.~Issac, N.~D. Ratliff, and
  D.~Fox, ``Closing the sim-to-real loop: Adapting simulation randomization
  with real world experience,'' in \emph{International Conference on Robotics
  and Automation, {ICRA} 2019, Montreal, QC, Canada, May 20-24, 2019}.\hskip
  1em plus 0.5em minus 0.4em\relax {IEEE}, 2019, pp. 8973--8979. [Online].
  Available: \url{https://doi.org/10.1109/ICRA.2019.8793789}
\BIBentrySTDinterwordspacing

\bibitem{du2021auto}
Y.~Du, O.~Watkins, T.~Darrell, P.~Abbeel, and D.~Pathak, ``Auto-tuned
  sim-to-real transfer,'' \emph{arXiv preprint arXiv:2104.07662}, 2021.

\bibitem{allevato2020tunenet}
A.~Allevato, E.~S. Short, M.~Pryor, and A.~Thomaz, ``Tunenet: One-shot residual
  tuning for system identification and sim-to-real robot task transfer,'' in
  \emph{Conference on Robot Learning}.\hskip 1em plus 0.5em minus 0.4em\relax
  PMLR, 2020, pp. 445--455.

\bibitem{Allevato:2020ui}
\BIBentryALTinterwordspacing
A.~D. Allevato, E.~Schaertl~Short, M.~Pryor, and A.~L. Thomaz, ``Iterative
  residual tuning for system identification and sim-to-real robot learning,''
  \emph{Autonomous Robots}, vol.~44, no.~7, pp. 1167--1182, 2020. [Online].
  Available: \url{https://doi.org/10.1007/s10514-020-09925-w}
\BIBentrySTDinterwordspacing

\bibitem{gat2017}
J.~P. Hanna and P.~Stone, ``Grounded action transformation for robot learning
  in simulation,'' in \emph{AAAI}.\hskip 1em plus 0.5em minus 0.4em\relax AAAI
  Press, 2017, pp. 4931--4932.

\bibitem{stocgat20}
S.~{Desai}, H.~{Karnan}, J.~P. {Hanna}, G.~{Warnell}, and a.~P.~{Stone},
  ``Stochastic grounded action transformation for robot learning in
  simulation,'' in \emph{2020 IEEE/RSJ International Conference on Intelligent
  Robots and Systems (IROS)}, 2020, pp. 6106--6111.

\bibitem{matas2018sim}
J.~Matas, S.~James, and A.~J. Davison, ``Sim-to-real reinforcement learning for
  deformable object manipulation,'' in \emph{Conference on Robot
  Learning}.\hskip 1em plus 0.5em minus 0.4em\relax PMLR, 2018, pp. 734--743.

\bibitem{evalutionary-direct-search}
\BIBentryALTinterwordspacing
V.~Heidrich-Meisner and C.~Igel, ``Hoeffding and bernstein races for selecting
  policies in evolutionary direct policy search,'' in \emph{Proceedings of the
  26th Annual International Conference on Machine Learning}, ser. ICML
  '09.\hskip 1em plus 0.5em minus 0.4em\relax New York, NY, USA: Association
  for Computing Machinery, 2009, pp. 401--408. [Online]. Available:
  \url{https://doi.org/10.1145/1553374.1553426}
\BIBentrySTDinterwordspacing

\bibitem{Rasmussen2004-wq}
C.~E. Rasmussen, ``Gaussian processes in machine learning,'' pp. 63--71, 2004.

\bibitem{nagabandi2018learning}
A.~Nagabandi, I.~Clavera, S.~Liu, R.~S. Fearing, P.~Abbeel, S.~Levine, and
  C.~Finn, ``Learning to adapt in dynamic, real-world environments through
  meta-reinforcement learning,'' in \emph{International Conference on Learning
  Representations}, 2018.

\bibitem{mendonca2020meta}
R.~Mendonca, X.~Geng, C.~Finn, and S.~Levine, ``Meta-reinforcement learning
  robust to distributional shift via model identification and experience
  relabeling,'' \emph{arXiv preprint arXiv:2006.07178}, 2020.

\bibitem{yu2020learning}
W.~Yu, J.~Tan, Y.~Bai, E.~Coumans, and S.~Ha, ``Learning fast adaptation with
  meta strategy optimization,'' \emph{IEEE Robotics and Automation Letters},
  vol.~5, no.~2, pp. 2950--2957, 2020.

\bibitem{perez2018film}
E.~Perez, F.~Strub, H.~De~Vries, V.~Dumoulin, and A.~Courville, ``Film: Visual
  reasoning with a general conditioning layer,'' in \emph{Proceedings of the
  AAAI Conference on Artificial Intelligence}, vol.~32, no.~1, 2018.

\bibitem{he2016deep}
K.~He, X.~Zhang, S.~Ren, and J.~Sun, ``Deep residual learning for image
  recognition,'' in \emph{Proceedings of the IEEE conference on computer vision
  and pattern recognition}, 2016, pp. 770--778.

\bibitem{gpy2014}
{GPy}, ``{GPy}: A gaussian process framework in python,''
  \url{http://github.com/SheffieldML/GPy}, since 2012.

\bibitem{emukit2019}
A.~Paleyes, M.~Pullin, M.~Mahsereci, N.~Lawrence, and J.~Gonz{\'a}lez,
  ``Emulation of physical processes with emukit,'' in \emph{Second Workshop on
  Machine Learning and the Physical Sciences, NeurIPS}, 2019.

\end{thebibliography}
